\documentclass{article}

\PassOptionsToPackage{round}{natbib}
\usepackage[preprint]{neurips_2021}

\usepackage[utf8]{inputenc} 
\usepackage[T1]{fontenc}    
\usepackage{url}            
\usepackage{booktabs}       
\usepackage{amsfonts}       
\usepackage{nicefrac}       
\usepackage{microtype}      
\usepackage{xcolor}         

\usepackage{wrapfig}
\usepackage{url}
\usepackage{breakurl}
\usepackage[breaklinks]{hyperref}
\usepackage{amsmath}
\usepackage{amsthm}
\usepackage{multirow}
\usepackage{adjustbox}
\usepackage{graphicx}
\usepackage{subcaption}
\usepackage[ruled,vlined]{algorithm2e}
\usepackage{enumitem}

\title{MAGI-X: Manifold-Constrained Gaussian Process Inference for Unknown System Dynamics}

\author{%
  Chaofan Huang\thanks{H. Milton Stewart School of Industrial and Systems Engineering, Georgia Institute of Technology, Atlanta, GA, 30332} \\ \texttt{chuang397@gatech.edu} 
  \And Simin Ma\footnotemark[1] \\ \texttt{sma318@gatech.edu} 
  \And Shihao Yang\footnotemark[1]\hspace{1.5mm}\thanks{To whom the correspondence should be addressed.} \\ \texttt{shihao.yang@isye.gatech.edu}
}

\begin{document}

\maketitle

\begin{abstract}
Ordinary differential equations (ODEs), commonly used to characterize the dynamic systems, are difficult to propose in closed-form for many complicated scientific applications, even with the help of domain expert. We propose a fast and accurate data-driven method, MAGI-X, to learn the \textit{unknown} dynamic from the observation data in a non-parametric fashion, without the need of any domain knowledge. Unlike the existing methods that mainly rely on the costly numerical integration, MAGI-X utilizes the powerful functional approximator of neural network to learn the unknown nonlinear dynamic within the MAnifold-constrained Gaussian process Inference (MAGI) framework that completely circumvents the numerical integration. Comparing against the state-of-the-art methods on three realistic examples, MAGI-X achieves competitive accuracy in both fitting and forecasting while only taking a fraction of computational time. Moreover, MAGI-X provides practical solution for the inference of partial observed systems, which no previous method is able to handle.
\end{abstract}

\section{Introduction}
\label{sec:introduction}

Systems of coupled ordinary differential equations (ODEs) are important tool for modeling the complex mechanism underlying many processes in science and engineering \citep{fitzhugh1961fn,nagumo1962fn,lotka1978lv,hirata2002hes1}. One successful application is in the infectious disease prediction, where the compartmental models have been the robust and interpretable tools for understanding the spread of many disease \citep{kermack1991sir}. However, given the overly complicated component interactions in a pandemic, the simple compartmental models fail short of capturing the underlying dynamic \citep{arik2020covid,karlen2020covid,yang2021covid,zou2020covid}. Careful ODE model building could yield better results, but it comes in the expense of domain expertise and many trials and errors in data fitting to work out an appropriate closed-form equation. Same challenge is also encountered in system biology and other scientific areas \citep{hirata2002hes1}. Thus, this raises the interest of efficiently learning the ODE structure purely from data when the true dynamics are partially or even completely unknown to the domain experts.

\paragraph{Problem formulation}
Consider the dynamic governed by the following system of ODEs,
\begin{equation}
  \label{eq:ode}
  \dot{x}(t) = \frac{dx(t)}{dt} = f(x(t),\theta,t), \; t\in[0,T],
\end{equation}
where $x(t)=(x_{1}(t),\ldots,x_{D}(t))\in\mathbb{R}^{D}$ is the output vector of the $D$-component dynamical systems at time $t\in[0,T]$, $\dot{x}(t)\in\mathbb{R}^{D}$ is the first order time derivative of $x(t)$, and $f$ is the $D$-dimensional vector-valued derivative function with parameters $\theta$ requiring estimation from the observation data. Computing the output at time $t$ requires the integration of $f$ over time, e.g., in the one-dimensional case, 
\begin{equation}
  \label{eq:ode_xt}
  x(t) = x_{0} + \int_{0}^{t}f(x(s),\theta,s) ds,
\end{equation}
where $x_{0} = x(0)$ is the initial state of the system. We assume that $f$ is \textit{completely unknown}, in which we have no prior information about either its parametric form nor the associated parameters. Moreover, we only observe some noisy output $y(\tau) = x(\tau) + \epsilon(\tau)$ at a set of discrete time points $\tau$ where $\epsilon$'s are the measurement errors. Given only the observations $y(\tau)$, our objective is to infer the \textit{black-box} derivative function $f$.

\paragraph{Related works}
The inference of the derivative function $f$ with known parametric form can date back to the \textit{classical (brute-force)} approach \citep{bard1974ode,vandomselaar1975ode,benson1979ode} that finds the parameters $\theta$ which minimizes the deviation between the observations and the system responses computed via \eqref{eq:ode_xt}, integration over time on $f$. This approach suffers the computational bottleneck from many evaluations of the costly integration. To circumvent the integration step, \citet{varah1982splineode} proposed an alternative approach known as the \textit{gradient constraint} that instead finds the parameter $\theta$ which minimizes the discrepancy in the two estimates of time derivatives: one from the smooth trajectory via de-noising the observations and the other from evaluating the derivative function $f$. Various data smoothing techniques have been considered, including splines \citep{varah1982splineode,ramsay2007splineode}, reproducing kernel Hilbert spaces \citep{gonzalez2014rkhsode,niu2016rkhsode}, and Gaussian process \citep{calderhead2009gpode,dondelinger2013gpode,wang2014gpode,macdonald2015gpode,wenk2019gpode}. However, it is not till recently that the introduction of \textit{MAnifold-constrained Gaussian process Inference} (MAGI) by \citet{yang2020magi} and \citet{wenk2020gpode} principally address the well known theoretical incompatibility of the Gaussian process based gradient constraint approach. MAGI not only shows great performance of inferring $\theta$, but more importantly, its runtime scales linearly in the number of components \citep{wenk2020gpode}.

In the recent decade, there has been growing interest of learning the derivative function $f$ that is \textit{completely unknown}, i.e., no parametric form of $f$ is accessible. Both Gaussian process \citep{heinonen2018odegp} and neural network \citep{chen2018odenn} have been proposed to model the \textit{black-box} derivative function $f$ under the \textit{classical} approach framework, but they suffer from the costly numerical integration that could be computational prohibitive for large system inference. Two prior works \citep{heinonen2014rkhsoderkhs,ridderbusch2020gpodegp} have attempted this problem under the two-step \textit{gradient constrained} framework where the data smoothing is performed only once and purely based on the observations, thus its performance is concerned by the quality of the smoothing being a good approximation of the true trajectory. Given the computational limitation and the performance instability observed in the existing techniques of unknown ODEs inference, we feel the urge of developing a computational efficient yet robust approach. To our best knowledge, no prior work has combined the sophisticated \textit{gradient constraint} approach such as MAGI with the learning of the unknown derivative function $f$.

\paragraph{Our contribution} 
Given the theoretical rigorousness, outstanding empirical performance, and computational efficiency of MAGI, we propose a novel algorithm that combines MAGI with the powerful approximator neural network for learning the unknown dynamic. MAGI-X elegantly addresses the following shortcomings in the existing literature of unknown ODE inference:

\begin{itemize}[noitemsep,topsep=0pt]
  \item By circumventing the costly integration step, MAGI-X only requires \textit{fraction} of the time (Table~\ref{tab:runtime}) to achieve comparable performance by state-of-the-arts based on the classical approach. Moreover, the runtime of MAGI-X scales linearly in the component dimension (Table~\ref{tab:large_system_runtime}), making it a preferable choice for large system inference.
  \item By using a feedback loop that incorporates information from $f$ in adapting the smoothing trajectory, MAGI-X improves the robustness over the two-step gradient constrained approach.
  \item By proposing a novel optimization algorithm that utilizes the generative mechanism inherently built in the Gaussian process for data augmentation, MAGI-X could avoid over-fitting of neural network even when the training data is sparse.
  \item By utilizing the Gaussian process, MAGI-X provides a principal probabilistic way to handle systems with sparse and partially observed data, which is common in real world applications.
\end{itemize}
Lastly, an PyTorch implementation of MAGI-X are provided on Github\footnote{\url{https://github.com/BillHuang01/MAGI-X}} for public use.

\section{MAGI: manifold-constrained Gaussian process inference}
\label{sec:magi}

\begin{wrapfigure}{r}{0.4\textwidth}
  \centering
  \includegraphics[width=0.36\textwidth]{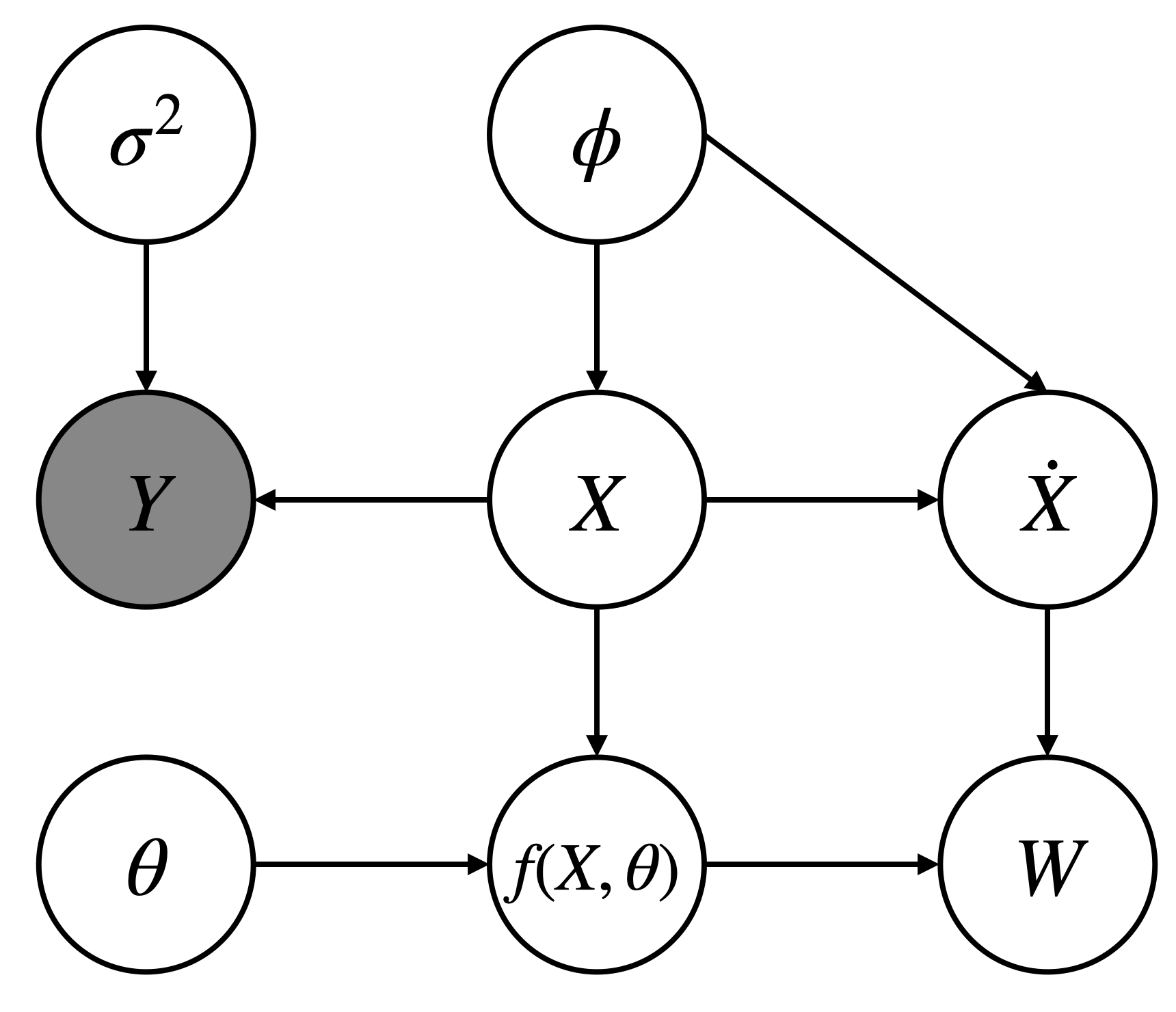}
  \caption{Graphical model of MAGI.}
  \label{fig:magi_pgm}
\end{wrapfigure}

Let us first present the MAGI \citep{wenk2020gpode,yang2020magi} for parameters inference of the derivative function $f$ with known parametric form. We focus on the framework of \citet{yang2020magi} with exact derivative estimates, and the corresponding graphical model is presented in Figure~\ref{fig:magi_pgm}. This section assumes the readers have some familiarity of Gaussian process (GP). A review of GP with its derivative computation is provided in the supplementary materials for the interested readers, and please refer to \citet{rusmassen2006gpml} for more details.

Following the Bayesian framework, we can view the $D$-component system response $x(t)$ as a realization of the stochastic processes $X(t) = (X_{1}(t),\ldots,X_{D}(t))$, and the derivative function parameters $\theta$ is a realization of the random variables $\Theta$. The goal is to compute the posterior distribution of $X(t)$ and $\Theta$, which requires first defining the prior and the data likelihood.

\paragraph{Prior} 
We assume a general prior $\pi$ on $\Theta$. For the stochastic process $X(t)$, we impose an independent GP prior on each component, that is, 
\begin{equation}
  \label{eq:gp_prior}
  X_{d}(t)\sim\mathcal{GP}(\mu_{d},\mathcal{K}_{d}),\; t\in[0,T],
\end{equation}
with mean function $\mu_{d}:\mathbb{R}\to\mathbb{R}$ and positive definite covariance function $\mathcal{K}_{d}:\mathbb{R}\times\mathbb{R}\to\mathbb{R}$ parameterized by hyperparameters $\phi_{d}$.

\paragraph{Likelihood}
Let us denote the observations $y(\tau) = (y_{1}(\tau_{1}),\ldots,y_{D}(\tau_{D}))$ where $\tau_{d} = (\tau_{d,1},\ldots,\tau_{d,N_{d}})$ is the set of $N_{d}$ observation time points for the $d$-th component. We allow different components to be observed at different sets of time points, but none of the component can be completely missing, that is $N_{d} > 0\; \forall d$. For simplicity, assuming the observation noise for the $d$-th component are i.i.d. zero mean Gaussian random variable with variance $\sigma^2_{d}$, then the observation likelihood is
\begin{equation}
  \label{eq:obs_likelihood}
  Y_{d}(\tau_{d})|X_{d}(\tau_{d}) = x_{d}(\tau_{d}) \sim \mathcal{N}(x_{d}(\tau_{d}), \sigma_{d}^2 I_{N_{d}}).
\end{equation}
Next, we introduce $W$ that quantifies the difference between the time derivative $\dot{X}(t)$ of GP and the ODE imposed gradient $f(X(t),\theta,t)$ with any given $\theta$,
\begin{equation}
  \label{eq:gm_w}
  W = \sup_{t\in[0,T], d\in\{1,\ldots,D\}}\bigg|\dot{X}_{d}(t) - \{f(X(t),\theta,t)\}_{d}\bigg|.
\end{equation}
Having $W\equiv 0$, the ideal case where the derivative function $f$ defined by $\theta$ is fully satisfied by $X(t)$, is to impose a \textit{manifold constraint} on the GPs that model $X(t)$. However, $W$ defined in \eqref{eq:gm_w} cannot be computed analytically since it is a supremum over an uncountable set. Thus, \citet{yang2020magi} propose an approximation $W_{\mathcal{T}}$ by finite discretization on $\mathcal{T} = (t_1,\ldots,t_n)\subset [0,T]$,
\begin{equation}
  \label{eq:gm_wa}
  W_{\mathcal{T}} = \max_{t\in \mathcal{T}, d\in\{1,\ldots,D\}}\bigg|\dot{X}_{d}(t) - \{f(X(t),\theta,t)\}_{d}\bigg|.
\end{equation}
It follows that a computable closed-form likelihood associated with the manifold constraint is 
\begin{equation}
  \label{eq:mc_likelihood}
    p\{W_{\mathcal{T}} = 0|X(\mathcal{T}) = x(\mathcal{T}), \Theta = \theta\} = p\{\dot{X}(\mathcal{T}) = f(x(\mathcal{T}),\theta,t)|X(\mathcal{T}) = x(\mathcal{T}), \Theta = \theta\}.
\end{equation}
Note that the time derivative of GP is also a GP with specific mean and covariance function (see supplementary materials for details), so \eqref{eq:mc_likelihood} is the p.d.f. of some multivariate Gaussian distribution.

\paragraph{Posterior}
It follows that the posterior distribution of $X(t)$ and $\Theta$ is
\begin{equation}
  \label{eq:posterior}
  \begin{aligned}
  & p\{\Theta=\theta,X(\mathcal{T})=x(\mathcal{T})|W_{\mathcal{T}}=0,Y(\tau)=y(\tau)\}  \\
  \propto \; & \pi_{\Theta}(\theta) \times \underbrace{p\{X(\mathcal{T}) = x(\mathcal{T})\}}_{\mbox{\eqref{eq:gp_prior}}} \times \underbrace{p\{Y(\tau)=y(\tau)|X(\tau)=x(\tau)\}}_{\mbox{\eqref{eq:obs_likelihood}}} \times \\
  & \underbrace{p\{W_{\mathcal{T}}=0|X(\mathcal{T})=x(\mathcal{T}),\Theta = \theta\}}_{\mbox{\eqref{eq:mc_likelihood}}}. 
  \end{aligned}
\end{equation}
Eqn. \eqref{eq:gp_prior}, \eqref{eq:obs_likelihood}, and \eqref{eq:mc_likelihood} are all p.d.f. of some multivariate normal distributions, so we can easily derive the log-posterior function (see supplementary materials).

\section{MAGI-X: MAGI for unknown ODEs dynamic}
\label{sec:magi-x}

Now consider the problem where the derivative function $f$ is \textit{completely unknown}. \citet{heinonen2018odegp} and \citet{ridderbusch2020gpodegp} both consider using vector-valued GP to model the unknown derivative function $f$. This suffer the computational complexity of $\mathcal{O}((DM)^{3})$ from computing the inverse of the covariance matrix, where $M$ is the number of ``inducing" locations, the set of spacing-filling points in the hypercube defined by the range of observation values, that the vector-valued GP interpolates. The usual rule of thumb is $M=10D$ \citep{loeppky2009size}, making vector-valued GP not suitable for large system learning. Thus, in this paper we consider an alternative class of powerful approximator, the neural networks, in which we can enjoy stochastic gradient descent (SGD) for more efficient learning. Moreover, the \textit{expressive power} of the neural network is not only well studied theoretically in the universal approximation theorem \citep{cybenko1989nn,hornik1989nn,barron1994nn}, but more importantly, it has shown great success in practice, making it a perfect choice for modelling $f$.

\paragraph{Key idea}
Though powerful, neural network is still inherently a parametric model parameterized by its connection weights and biases. By using a neural network to model the unknown derivative function $f$, we can employ MAGI discussed in Section~\ref{sec:magi} to efficiently identify the network parameters $\theta$ such that the learned neural network well approximates $f$, leading to our proposed method MAGI-X. However, finding an efficient yet robust optimization procedure to solve for $\theta$ is not trivial, which we will discuss in details later in this section.

\paragraph{Prior}
For the network parameters $\theta$, the natural prior choice is a non-informative prior, that is $\pi(\theta)\propto 1$. The noise variance $\sigma^2 = (\sigma_{1}^2,\ldots,\sigma_{D}^2)$ is generally not known apriori, so a non-informative prior can also be assigned, that is $p(\sigma^2)\propto 1$. 

\paragraph{Objective function}
From Section~\ref{sec:magi}, we show that the random variable $W$ associated with the \textit{manifold constraint} requires an approximation $W_{\mathcal{T}}$ on finite set of discretized time points $\mathcal{T}$. $W_{\mathcal{T}}\to W$ monotonically as $\mathcal{T}$ becomes dense \citep{yang2020magi}. Thus, we suggest to choose $\mathcal{T}$ that is a much finer discretization than the observation time points, where $|\mathcal{T}| > N_{d}\; \forall d$, but this leads to stronger emphasis on the GP prior \eqref{eq:gp_prior} and the manifold constraint likelihood \eqref{eq:mc_likelihood} in the posterior function \eqref{eq:posterior}, especially if we allow the cardinality of $\mathcal{T}$ to be arbitrarily large for a precise approximation. To balance out the influence from the observations and the discretization points, we introduce some tempering parameters on the observation likelihood \eqref{eq:obs_likelihood}, leading to the objective function,
\begin{equation}
  \label{eq:objective}
  \begin{small}
  \begin{aligned}
  \arg\max_{\theta,x(\mathcal{T}),\sigma^2} &\underbrace{\log p\{X(\mathcal{T}) = x(\mathcal{T})\}}_{\mbox{\eqref{eq:gp_prior}}} + \underbrace{\log p\{W_{\mathcal{T}}=0|X(\mathcal{T})=x(\mathcal{T}),\Theta = \theta\}}_{\mbox{\eqref{eq:mc_likelihood}}} + \\
  &\underbrace{\sum_{d=1}^{D} \frac{|\mathcal{T}|}{N_{d}}\log p\{Y_{d}(\tau)=y_{d}(\tau)|X_{d}(\tau)=x_{d}(\tau),\sigma_{d}^2\}}_{\mbox{\eqref{eq:obs_likelihood}}},
  \end{aligned}
  \end{small}
\end{equation}
that we can solve for the Maximum a Posteriori (MAP) estimation of the tempering adjusted posterior. The MAP estimation for $\theta$ is sufficient given that the uncertainty quantification of network weights and biases does not have good physical interpretation.

\begin{algorithm}[t!]
\SetAlgoLined
  \textbf{Input:} (i) observations $y(\tau) = \{y_{d}(\tau_{d})\}_{d=1}^{D}$, (ii) the discretized time points $\mathcal{T}=(t_{1},\ldots,t_{n})$, and (iii) the neural network architecture with its associated parameters $\theta$. \\
  \vspace{2mm}
  \textit{Initialization}: \\
  $\bullet$ time standardization (see supplementary material for details). \\
  \For{$d = 1,\ldots,D$}{
    $\bullet$ fit GP regression on $y_{d}(\tau_{d})$ to identify the hyperparameters $\phi_{d}$ and noise variance $\sigma_{d}^2$. \\
    $\bullet$ initialize $x_{d}(\mathcal{T})$ using the predictive mean of the trained GP evaluated at $\mathcal{T}$.
  }
  $\bullet$ initialize $\theta$ by optimize \eqref{eq:objective} as a function of $\theta$ only by fixing $x(\mathcal{T})$ and $\{\sigma_{d}^2\}_{d=1}^{D}$ at initial values. \\
  \vspace{2mm}
  \textit{Optimization} (block-wise update alternating between $\theta$, $x(\mathcal{T})$, and $\{\sigma_{d}^2\}_{d=1}^{D}$): \\
  \For{$l = 1,\ldots,L$}{
    $\bullet$ gradient ascent on $\theta$ with learning rate $\eta^{(\theta)}_{l} = 0.005 l^{-0.6}$. \\
    $\bullet$ gradient ascent on $x(\mathcal{T})$ with learning rate $\eta^{(x)}_{l} = 0.05 (500+l)^{-0.6}$. \\
    $\bullet$ closed form update for $\{\sigma_{d}^2\}_{d=1}^{D}$. 
  }
  \vspace{-\topsep}
  \vspace{2mm}
  \textbf{Return:} optimized $\theta$, $x(\mathcal{T})$, and $\{\sigma_{d}^2\}_{d=1}^{D}$.  \\
 \caption{MAGI-X.}
 \label{algo:magi-x}
\end{algorithm}

\paragraph{Optimization procedure}
There are two levels of functional approximation in MAGI-X: the manifold-constrained GP approximation of the true trajectory and the neural network approximation of the derivative function $f$ that yields the constraint, making the optimization problem in \eqref{eq:objective} difficult to solve. Thus, a good robust initialization is required. We start with a time standardization to make sure that MAGI-X is numerically robust for problems across different time ranges. Following the initialization procedure in \citet{yang2020magi}, we fit independent GP regressions on the observations of each component $y_{d}(\tau_{d})$, and then initialize $x_{d}(\mathcal{T})$ by the predictive mean of the GP regression at $\mathcal{T}$. The GP hyperparameters $\{\phi_{d}\}_{d=1}^{D}$ and the noise variance $\{\sigma_{d}^2\}_{d=1}^{D}$ are obtained by the empirical Bayes method. To initialize $\theta$, we optimize over \eqref{eq:objective} as a function of only $\theta$ while fixing $x(\mathcal{T})$ and $\{\sigma_{d}^2\}_{d=1}^{D}$ at their initialized values. The initialization procedure is equivalent to the two-step approach employed in \citet{ridderbusch2020gpodegp} but with neural network for modeling $f$. However, in practice, we see that the learned dynamic from initialization generalizes poorly to unseen data, leading to bad forecasting performance. Thus, subsequent \textit{feedback-loop} optimization alternating between $x(\mathcal{T})$ and $\theta$ is required. For computational efficiency, the hyperparameters $\{\phi_{d}\}_{d=1}^{D}$ are kept fixed after initialization.

In the subsequent optimization stage, we consider the block-wise update alternating between $\theta$, $x(\mathcal{T})$, and $\sigma^2$. Prior to each update of $\theta$, we perform a coordinate ascent on $x(\mathcal{T})$, yielding new input/output pairs $\{x(\mathcal{T}),\dot{x}(\mathcal{T})\}$ for computing the gradient of $\theta$. In other words, thanks to the Bayesian GP framework, this step naturally yields a implicit ``generative" mechanism that can simulate infinite reasonable yet slightly noisy input/output pairs for training the network parameters $\theta$. Given that the input/output pairs can be viewed as noisy samples simulated from the distribution of the groundtruth trajectory, we employ a polynomially decayed learning rate $\eta_{l}=a(b+l)^{-\gamma}$ at the $l$-th iteration to ensure convergence. Moreover, the new set of unseen input/output pair can also serve as a validation set to encourage generalization, which is the key to avoid over-fitting of neural network when the observations are sparse. Algorithm~\ref{algo:magi-x} details the optimization procedure of MAGI-X, and some minor implementation details for numerical stability are discussed in the supplementary materials.

\section{Simulation results}
\label{sec:simulation_results}

In this section we first study the performance of MAGI-X on three real world systems: the FitzHugh-Nagumo (FN) in \citet{fitzhugh1961fn} and \citet{nagumo1962fn}, the Lotka-Volterra (LV) in \citet{lotka1978lv}, and the Hes1 in \citet{hirata2002hes1}. Since LV and Hes1 are strictly positive systems, we consider both systems after the log-transformation. Next, we also apply MAGI-X on a synthetic high dimensional example to demonstrate its runtime scaled linearly in the number of components. The functional forms of the aforementioned systems are provided in the supplementary material. Note that our objective is to learn the \textit{unknown} dynamic, so only noisy observations are fed to MAGI-X for the inference. The information on the parametric forms are not available for MAGI-X.

\paragraph{Data generation} 
All groundtruth data are simulated by numerical integration. Since FN, LV, and Hes1 are oscillators, we obtain the true trajectory on 321 equal spaced-out time points that approximately form four to five cycles (see dashed lines in Figure~\ref{fig:trajectory}). To generate the noisy observations $y(\tau)$, we use \textit{one fourth subset} time points of the first 161 data points with added i.i.d. Gaussian random noise, i.e., there are only 41 observations available for each component. We refer to the first 161 points as \textit{fitting} phase. The rest 160 points are left for forecasting performance evaluation, and thus we refer to as \textit{forecasting} phase.

\paragraph{Evaluation metric}
We employ trajectory root mean squared error (RMSE) as the evaluation metric to measure the quality of the model in recovering the true system. We compute the trajectory RMSE for (i) different components, (ii) different phases: fitting vs. forecasting, and (iii) different types: inferred vs. reconstructed. The \textit{inferred} trajectory refers to the $x(\mathcal{T})$ from the output of MAGI-X (Algorithm~\ref{algo:magi-x}) plus the forecast trajectory computed via numerical integration with initial condition being the last point of $x(\mathcal{T})$. The \textit{reconstructed} trajectory refers to the trajectory computed via numerical integration with initial condition being the inferred $x_{0}$, i.e., the first point of inferred $x(\mathcal{T})$.  We emphasize that numerical integration is only used for the evaluation and forecast in MAGI-X, and during the in-sample fitting, \textit{no numerical integration is ever needed}.

\paragraph{MAGI-X setting}
First is regarding to the discretization $\mathcal{T}$. Given that $y(\tau)$ are from the fitting phase only, we use the same first 161 equal spaced-out time points for $\mathcal{T} = \{t_{1},t_{2},\ldots,t_{161}\}$ that is four times denser than the observation time points $\tau$. This choice of $\mathcal{T}$ also allows for easy computation of the \textit{inferred} trajectory RMSE. In general, we suggest at least four times finer discretization than the number of available observations. Next, for the GP covariance function, we employ the Mat\'{e}rn kernel with degree of freedom 2.01 that is suggested in \citet{yang2020magi} to ensure the kernel is only twice differentiable. In practice, the Mat\'{e}rn kernel is generally more robust than the commonly used Radial Basis Function (RBF) kernel for our problem of interest, as the covariance matrix of the GP derivative would be too smooth if RBF kernel is used. For the neural network, we use ReLU activation function. In this section we only present the result for the neural network with a single hidden layer of 512 nodes that yields the best performance, but we provide the comparison of different network structures in the supplementary materials for the interested readers. We run MAGI-X for 2{,}500 iterations.

\paragraph{Benchmark models}
We compare MAGI-X to the two state-of-the-arts, NPODE in \citet{heinonen2018odegp} and Neural ODE in \citet{chen2018odenn}, that both rely on numerical integration for unknown dynamic inference. Implementation provided by the respective authors on Github are used. For the NPODE, we use the default GP setting and run it for 500 iterations as suggested. For the Neural ODE, we employ the same network structure, a single hidden layer of 512 nodes, utilized in MAGI-X for a fair comparison. The $\mathcal{O}(1)$ memory adjoint method is used for the backpropagation. However, due to its long training time (Table~\ref{tab:runtime}), we only run Neural ODE for 500 iterations. For both NPODE and Neural ODE, the \textit{inferred} trajectory is identical to the \textit{reconstructed} trajectory.

\subsection{Fully Observed System}
\label{subsec:fully_observed_system}

\begin{figure}[h!]
  \centering
  \begin{subfigure}{0.32\textwidth}
    \centering
    \includegraphics[width=0.95\textwidth]{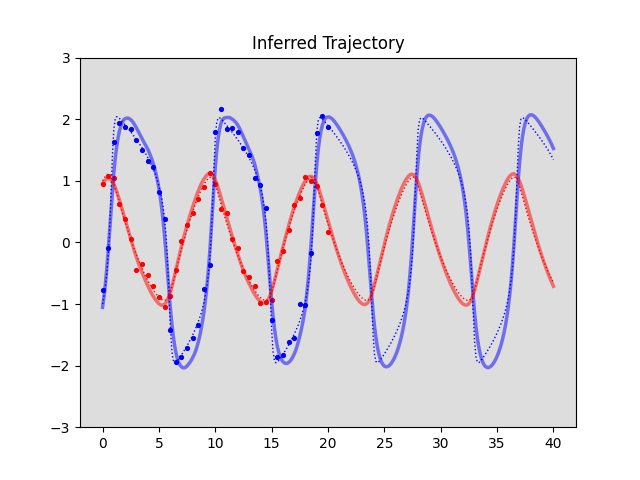}
    \caption{FN}
  \end{subfigure}
  \begin{subfigure}{0.32\textwidth}
    \centering
    \includegraphics[width=0.95\textwidth]{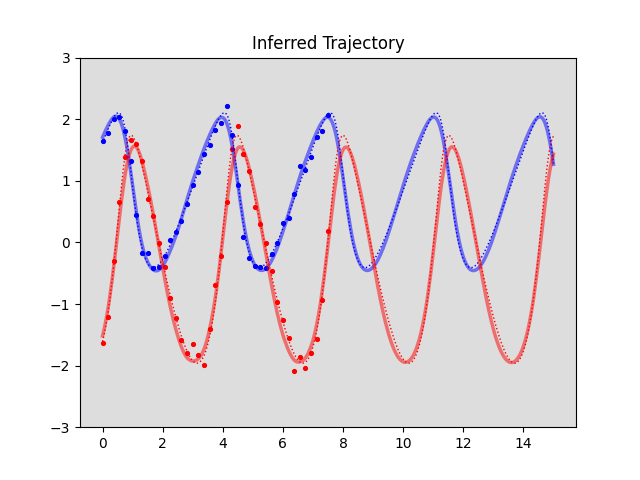}
    \caption{LV (log-transformed)}
  \end{subfigure}
  \begin{subfigure}{0.32\textwidth}
    \centering
    \includegraphics[width=0.95\textwidth]{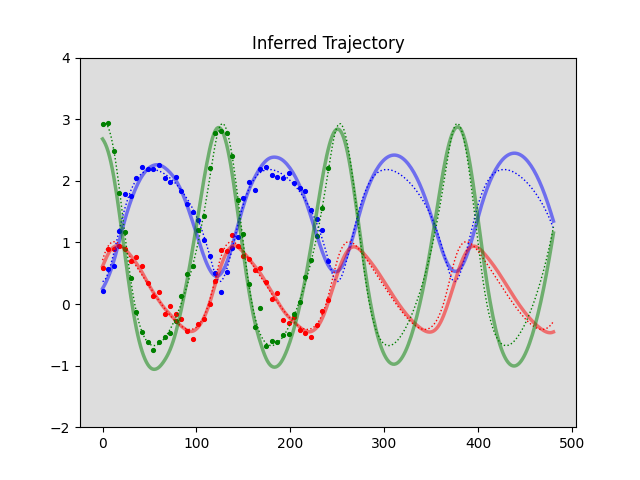}
    \caption{Hes1 (log-transformed)}
  \end{subfigure}
  \caption{Comparison of the inferred trajectory (solid line) to the groundtruth trajectory (dotted lines) after applying MAGI-X (Algorithm~\ref{algo:magi-x}) on the 41-point fully observed data (circles).}
  \label{fig:trajectory}
\end{figure}

\begin{table}[h!]
  \caption{Means and standard deviations of trajectory RMSEs over 100 runs with full observations. The best results are highlighted in red boldface.}
  \label{tab:trajectory_rmse}
  \centering
  \resizebox{\textwidth}{!}{%
  \begin{tabular}{lllccccccc}
    \toprule
    & & & \multicolumn{2}{c}{FN} & \multicolumn{2}{c}{LV} & \multicolumn{3}{c}{Hes1} \\
    \cmidrule(lr){4-5}\cmidrule(lr){6-7}\cmidrule(lr){8-10}
    Phase & Model & Trajectory & $x_1$ & $x_2$ & $x_1$ & $x_2$ & $x_1$ & $x_2$ & $x_3$ \\
    \midrule
    \multirow{4}{5em}{Fitting} & NPODE & Reconstructed & $0.25 \pm 0.14$ & $0.06 \pm 0.04$ & \textcolor{red}{$\boldsymbol{0.04 \pm 0.01}$} & \textcolor{red}{$\boldsymbol{0.05 \pm 0.01}$} & $1.46 \pm 1.37$ & $1.25 \pm 1.17$ & $2.07 \pm 1.35$ \\
    \cmidrule{2-10}
    & Neural ODE (512) & Reconstructed & $1.11 \pm 0.34$ & $0.48 \pm 0.17$ & $0.39 \pm 0.18$ & $0.65 \pm 0.41$ & - & - & -  \\
    \cmidrule{2-10}
    & \multirow{2}{8em}{MAGI-X (512)} & Reconstructed & \textcolor{red}{$\boldsymbol{0.21 \pm 0.02}$} & \textcolor{red}{$\boldsymbol{0.05 \pm 0.01}$} & $0.06 \pm 0.01$ & $0.07 \pm 0.03$ & $0.12 \pm 0.05$ & $0.06 \pm 0.02$ & $0.13 \pm 0.03$ \\
    & & Inferred & \textcolor{red}{$\boldsymbol{0.21 \pm 0.02}$} & \textcolor{red}{$\boldsymbol{0.05 \pm 0.01}$} & $0.05 \pm 0.01$ & $0.06 \pm 0.01$ & \textcolor{red}{$\boldsymbol{0.11 \pm 0.05}$} & \textcolor{red}{$\boldsymbol{0.06 \pm 0.01}$} & \textcolor{red}{$\boldsymbol{0.12 \pm 0.04}$} \\
    
    \midrule
    
    \multirow{4}{5em}{Forecasting} & NPODE & Reconstructed & $0.32 \pm 0.25$ & $0.11 \pm 0.11$ & \textcolor{red}{$\boldsymbol{0.07 \pm 0.04}$} & \textcolor{red}{$\boldsymbol{0.09 \pm 0.05}$} & $1.58 \pm 1.48$ & $1.39 \pm 1.30$ & $2.21 \pm 1.39$ \\
    \cmidrule{2-10}
    & Neural ODE (512) & Reconstructed & $1.40 \pm 0.34$ & $0.63 \pm 0.18$ & $0.59 \pm 0.20$ & $0.91 \pm 0.36$ & - & - & -  \\
    \cmidrule{2-10}
    & \multirow{2}{8em}{MAGI-X (512)} & Reconstructed & \textcolor{red}{$\boldsymbol{0.21 \pm 0.03}$} & \textcolor{red}{$\boldsymbol{0.06 \pm 0.02}$} & $0.12 \pm 0.10$ & $0.18 \pm 0.24$ & $0.25 \pm 0.14$ & $0.13 \pm 0.07$ & $0.30 \pm 0.17$ \\
    & & Inferred & \textcolor{red}{$\boldsymbol{0.21 \pm 0.03}$} & \textcolor{red}{$\boldsymbol{0.06 \pm 0.02}$} & $0.09 \pm 0.04$ & $0.12 \pm 0.06$ & \textcolor{red}{$\boldsymbol{0.24 \pm 0.14}$} & \textcolor{red}{$\boldsymbol{0.13 \pm 0.06}$} & \textcolor{red}{$\boldsymbol{0.30 \pm 0.16}$} \\
    \bottomrule
  \end{tabular}
  }
\end{table}

\begin{table}[t!]
  \caption{Means and standard deviations of computational time (seconds) over 100 simulated runs. All experiments are ran under the same CPU setting.}
  \label{tab:runtime}
  \centering
  \begin{tabular}{lccc}
    \toprule
    Model & FN & LV & Hes1 \\
    \midrule
    NPODE & $42.82 \pm 3.78$ & $42.78 \pm 3.79$ & $102.88 \pm 28.16$ \\
    Neural ODE (512) & $570.08 \pm 35.15$ & $511.52 \pm 67.83$ & - \\
    MAGI-X (512) & $16.64 \pm 0.75$ & $17.05 \pm 0.87$ & $21.44 \pm 3.51$ \\
    \bottomrule
  \end{tabular}
\end{table}

We first consider the simple case where all components share the same set of observation time points, i.e., $\tau_{1} = \cdots = \tau_{D}$. In other words, at any particular observation time point, we have data for all components in the system, which is the general setting considered in existing literature. We simulate 41 noisy observations at one-out-of-every-four points from the fitting phase (first 161 points), i.e., $\tau_{d} = \{t_{1},t_{5},\ldots,t_{157},t_{161}\}\;\forall d$. We use the same noise level in \citet{heinonen2018odegp} with variance $\sigma_{d}^2 = 0.1^2\; \forall d$. The circles in Figure~\ref{fig:trajectory} shows one set of the 41-point synthetic noise-contaminated observations from the FN, LV, and Hes1 systems. Figure~\ref{fig:trajectory} shows the inferred trajectory (solid line) learned by MAGI-X (Algorithm~\ref{algo:magi-x}), and we can see that it recovers the true trajectory (dashed line) well visually for the FN and LV system. For the Hes1 example, MAGI-X is able to capture the periodic pattern but slightly over-estimates the peaks of some components. However, we see that the inferred trajectory over-smooth the true dynamic in all three examples, which is the inherent limitation of GP approximation due to the bias introduced by the prior.

To further assess the performance of MAGI-X and compare to the state-of-the-art methods, we simulate 100 independent datasets each with 41 noisy observations under the aforementioned setting but with different random seeds. Table~\ref{tab:trajectory_rmse} shows the mean and standard deviation trajectory RMSEs of NPODE, Neural ODE, and MAGI-X on the three benchmark systems. For the FN example, MAGI-X outperforms the NPODE and Neural ODE significantly in both fitting and forecasting phase. For the LV example, the performance of the inferred trajectory by MAGI-X is comparable to that of the NPODE, but the reconstructed trajectory underperforms in the forecasting phase with large standard deviation in $x_2$ (0.24). This is because one of the trajectory diverges away from the observation domain, and after removing this outlier, the means and standard deviation of RMSE in reconstructed $x_2$ is $0.16\pm 0.12$, which is only slightly worse than NPODE. We will discuss more about this numerical optimization issue and possible solution in the supplementary materials. For the Hes1 example, the performance of NPODE is disastrous, which could be due to the numerical instability by the large time scale (in 100s) of Hes1, where the time scale of FN and LV is in 10s. Because of the time standardization in preprocessing step of Algorithm~\ref{algo:magi-x}, MAGI-X does not suffer such instability and yields comparable results to the FN and LV examples. The poor performance of the Neural ODE might be due to the fact that 500 iterations are not enough for convergence, but it already incurs a much larger computational cost than the NPODE and MAGI-X (Table~\ref{tab:runtime}). Due to limited computational budgets, we do not run Neural ODE on the Hes1 example. Furthermore, Table~\ref{tab:runtime} shows that under the same CPU setting, MAGI-X enjoys more than two times faster runtime than the NPODE on the two-component systems, and this computational saving is more significant on the three-component Hes1 system: MAGI-X only requires \textit{one fifth} of the runtime by NPODE. This clearly shows the computational advantage of MAGI-X, not to mention that the performance of MAGI-X is comparable or even better than the two state-of-the-art methods. Moreover, the success of MAGI-X demonstrate that the inherently built-in ``generative" mechanism in the proposed optimization procedure could help avoid over-fitting of neural network under sparse training samples setting (82 observations for the two-component system and 123 observations for the three-component system).

\subsection{Partially Observed System}
\label{subsec:partially_observed_system}

\begin{table}[t!]
  \caption{Means and standard deviations of trajectory RMSEs over 100 runs with partial observations.}
  \label{tab:trajectory_rmse_partial}
  \centering
  \resizebox{\textwidth}{!}{%
  \begin{tabular}{lllccccccc}
    \toprule
    & & & \multicolumn{2}{c}{FN} & \multicolumn{2}{c}{LV} & \multicolumn{3}{c}{Hes1} \\
    \cmidrule(lr){4-5}\cmidrule(lr){6-7}\cmidrule(lr){8-10}
    Phase & Model & Trajectory & $x_1$ & $x_2$ & $x_1$ & $x_2$ & $x_1$ & $x_2$ & $x_3$ \\
    \midrule
    \multirow{2}{5em}{Fitting} & \multirow{2}{6em}{MAGI-X (512)} & Reconstructed & $0.21 \pm 0.02$ & $0.06 \pm 0.02$ & $0.06 \pm 0.02$ & $0.07 \pm 0.03$ & $0.12 \pm 0.04$ & $0.07 \pm 0.02$ & $0.15 \pm 0.04$  \\
    & & Inferred & $0.21 \pm 0.02$ & $0.06 \pm 0.01$ & $0.06 \pm 0.02$ & $0.06 \pm 0.01$ & $0.12 \pm 0.04$ & $0.07 \pm 0.02$ & $0.14 \pm 0.04$ \\
    
    \midrule
    \multirow{2}{5em}{Forecasting} & \multirow{2}{6em}{MAGI-X (512)} & Reconstructed & $0.21 \pm 0.04$ & $0.06 \pm 0.03$ & $0.13 \pm 0.10$ & $0.19 \pm 0.15$ & $0.29 \pm 0.14$ & $0.17 \pm 0.07$ & $0.39 \pm 0.18$  \\
    & & Inferred & $0.21 \pm 0.03$ & $0.06 \pm 0.03$ & $0.11 \pm 0.07$ & $0.15 \pm 0.09$ & $0.29 \pm 0.13$ & $0.17 \pm 0.07$ & $0.39 \pm 0.17$ \\
    \bottomrule
  \end{tabular}
  }
\end{table}

On top of its significant computational advantage, MAGI-X is the first unknown ODEs inference method that provides a practical and complete solution to handle systems with asynchronous observation times, where $\tau_{1} \neq \cdots \neq \tau_{D}$. This is commonly seen in practice, since sensor failure or human error could possibly lead to missing observations of some components but not necessary the whole system at a particular time point. Thanks to the Bayesian GP framework, MAGI-X provides a principal probabilistic framework for the missing data imputation, which naturally takes care of the partial observation scenario without any modification. However, MAGI-X cannot recover a component that is completely unobserved, i.e. $\tau_{d} = \emptyset$, as it inevitably suffers from model identifiability issue.

To exam the robustness of MAGI-X against the partially observed systems, we consider simulating the noisy observations in the following extreme setting. For the two-component system FN and LV, we only allow the observation of one component at each time point. For a fair comparison to the results of the fully observed system, we keep the number of observations to be about 41, and thus letting $\tau_{1} = \{t_{1},t_{5},\ldots,t_{161}\}$ and $\tau_{2} = \{t_{3},t_{7},\ldots,t_{159}\}$. Again, i.i.d. Gaussian random noise with variance $\sigma_{d}^2 = 0.1^2\; \forall d$ is applied to contaminate the groundtruth trajectory value. We simulate 100 independent partially observed datasets under the above setting, and Table~\ref{tab:trajectory_rmse_partial} shows the means and standard deviations of the trajectory RMSEs. We can see that on the FN and LV systems, the results are comparable to the setting of fully observed system in Table~\ref{tab:trajectory_rmse}, showing that MAGI-X still performs well even if only partial observations are available.

For the Hes1 system, let us consider a even more challenging setting with fewer observations. We let $\tau_{1} = \{t_{5},t_{9},t_{17},t_{21},\ldots,t_{161}\}$, $\tau_{2} = \{t_{1},t_{9},t_{13},t_{21},\ldots,t_{157}\}$, and $\tau_{3} = \{t_{1},t_{5},t_{13},t_{17},\ldots,t_{161}\}$. Here, (i) only two out of three components are observable at a particular time, (ii) the observation times are no longer equal spaced-out, (iii) only about 28 observations are available for each component, which is one third less than the 41 observations considered previously. We again simulate 100 independent datasets each with different i.i.d. $\sigma_{d}^2 = 0.1^2$ variance Gaussian random noise. Table~\ref{tab:trajectory_rmse_partial} shows the means and standard deviations of the resulted trajectory RMSEs. Compared to the performance of the 41-point fully observed data in Table~\ref{tab:trajectory_rmse}, we can see that it is about $\sqrt{41/28}$ worse, well within the expected accuracy one can possibly hope for under this challenging example, further demonstrating the robustness of the MAGI-X on the partially observed systems that no existing method is able to handle.

\subsection{Linear Scaling in Component Dimension}
\label{subsec:linear_scaling_in_component_dimension}

\begin{table}[t!]
  \caption{Means and standard deviations of computational time (seconds) on the synthetic large system example over 100 runs. All experiments are ran under the same CPU setting.}
  \label{tab:large_system_runtime}
  \centering
  \begin{tabular}{cc}
    \toprule
    Component Dimension & Computational Time \\
    \midrule
    10 & $72.26 \pm 3.43$ \\
    20 & $135.33 \pm 2.57$ \\
    40 & $272.65 \pm 3.77$ \\
    \bottomrule
  \end{tabular}
\end{table}

For the ODEs parameter inference where the derivative function $f$ has known parametric form, \citet{wenk2020gpode} have demonstrated that the runtime of the \textit{gradient constrained} approach scales linearly in the number of components. We also want to show that such linear scaling computational advantage can be preserved in MAGI-X, in which the parametric form of $f$ is completely \textit{unknown}. For illustration, we apply MAGI-X on a synthetic Hamiltonian system ODEs dynamic with 10, 20, and 40 components. Following the same data generation procedure, we again simulate 41-point noisy observations for each component by adding i.i.d. $0.1^2$ variance Gaussian random noise. 

Table~\ref{tab:large_system_runtime} shows that the experimental runtime of MAGI-X increases linearly in the number of components. The linear scaling of MAGI-X is achieved by (i) its \textit{gradient constrained} framework that circumvents the costly numerical integration and (ii) the use of neural network for approximating the derivative function $f$. If vector-valued GP is employed for learning $f$ such as in \citet{heinonen2018odegp} and \citet{ridderbusch2020gpodegp}, the linear scaling would not be feasible due to the cubic complexity required for computing the inverse of covariance matrix. More surprisingly, the runtime of MAGI-X on the 40-component system is only about 270 seconds (Table~\ref{tab:large_system_runtime}), while Neural ODE takes more than 500 seconds for training a two-component system (Table~\ref{tab:runtime}). Again, this demonstrate the remarkable computational efficiency of MAGI-X on large system inference over the existing state-of-the-art methods that is built upon numerical integration, i.e., the \textit{classical} approach for ODEs dynamic inference. Detail experimental results on this synthetic large system example are provided in the supplementary materials.

\section{Discussion}
\label{sec:discussion}

In this paper we propose MAGI-X, an elegant extension of the MAnifold-constrained Gaussian process Inference (MAGI) for learning \textit{unknown} ODEs system, where the associated derivative function $f$ does not have known parametric form. To model the black-box $f$, we utilize the neural network, an powerful class of universal approximator in both theory and practice. From the numerical results presented in Section~\ref{sec:simulation_results}, we see that MAGI-X not only performs comparably to the state-of-the-art methods, but more importantly, it enjoys drastic reduction in runtime by bypassing the costly numerical integration. Moreover, a synthetic large system example is also included to demonstrate its linear scaling computational time in the component dimensions. The computational advantage of MAGI-X makes it a feasible choice for large system inference, even on a personal laptop. Last, the Bayesian GP framework employed in MAGI-X provides a principal way for handling partial observed systems that is neglected in the existing literature.

Though MAGI-X might be seen as a simple extension of MAGI by replacing the parametric derivative function $f$ with a neural network, this replacement causes many difficulties in finding a robust optimization procedure, as MAGI-X now needs to deal with two levels of functional approximation: (i) the underlying trajectory $X(t)$ by manifold-constrained GP, and (ii) the unknown derivative function $f$ by neural network. In this paper we propose a robust optimization procedure based on block-wise update alternating between the inferred trajectory $x(\mathcal{T})$ and the network parameters $\theta$: $x(\mathcal{T})$ is first updated by a small perturbation, and then $\theta$ is updated to better fit the perturbed $x(\mathcal{T})$. Consequently, $\theta$ is trained with many probable $x(\mathcal{T})$ generated during optimization, thus avoids over-fitting issue from sparse observation $y(\tau)$. This robust optimization essentially enforces the MAGI-X feedback loop with GP. From Figure~\ref{fig:trajectory} we can see that the derivative function $f$ learned by neural network is able to recover the periodic pattern without enforcing periodic GP approximation of the underlying dynamic. This exhibit the great extrapolation property of our proposed procedure, otherwise we would likely observe a divergent trend in the forecasting phase.

Though we have achieve great success in using neural networks for good approximation of the unknown derivative function $f$, this is only useful for forecasting, as the neural network is still black-box model that cannot provide any interpretations. One possible follow-up research direction is to recover the structure (parametric form) that is implicitly learned by the neural network. The parametric form could then addresses the two major drawbacks of any powerful nonlinear approximator: (i) lack of interpretability, and (ii) poor extrapolation to unseen input. Moreover, this would provide an efficient end-to-end purely data-driven approach for identifying the parametric form that could best capture the unknown dynamic of interest without the need of domain knowledge. 

Yet, a more realistic setting is when the system of ODEs is partially known, i.e., there exist some investigated parametric forms but all fail to capture the true dynamic underlying the noisy observations. Previous approaches have tried to address the discrepancy by using Kennedy-O'Hagan framework \citep{kennedy2001calibration} on the trajectory level. With the recent development of learning unknown ODEs, we can easily adapt the Kennedy-O'Hagan framework on the derivative function level to correct the discrepancy between the existing inaccurate parametric form and the oracle but unknown derivative function $f$, which is another interesting future direction. 

\medskip

\bibliography{references}

\newpage

\setcounter{table}{0}
\renewcommand{\thetable}{S\arabic{table}}%
\setcounter{figure}{0}
\renewcommand{\thefigure}{S\arabic{figure}}%
\setcounter{section}{0}

\appendix

\section{Gaussian process}
\label{appendix:gaussian_process}

We introduce the scalar-input scalar-output Gaussian process that is concerned in this paper. Following the definition in \citet{rusmassen2006gpml}, a Gaussian Process is a collection of random variables such that any finite number of which have a joint multivariate Gaussian distribution, denoted by 
\begin{equation}
  \label{eq:gp1}
  X(t) \sim \mathcal{GP}(\mu,\mathcal{K}_{\phi}),\; t\in\mathbb{R},
\end{equation}
where $\mu:\mathbb{R}\to\mathbb{R}$ is the mean function and $\mathcal{K}_{\phi}:\mathbb{R}\times\mathbb{R}\to\mathbb{R}$ is a positive definite covariance function parameterized by some hyperparameter $\phi$. For any finite set of time points $\mathcal{T} = (t_{1},\ldots,t_{n})$, we have 
\begin{equation}
  \label{eq:gp2}
  X(\mathcal{T})\sim\mathcal{N}(\mu(\mathcal{T}),\mathcal{K}_{\phi}(\mathcal{T},\mathcal{T})).
\end{equation}
 
It is typical that we only have noisy observations of the function values, i.e., we observe $Y(t_i) = X(t_{i}) + \epsilon_{i}$ where we assume additive i.i.d. Gaussian random noise $\epsilon_{i}\sim\mathcal{N}(0,\sigma^2)$. It follows that 
\begin{equation}
  \label{eq:gp3}
  Y(\mathcal{T}) \sim \mathcal{N}(\mu(\mathcal{T}), \mathcal{K}_{\phi}(\mathcal{T},\mathcal{T}) + \sigma^{2} I_n).
\end{equation}
Conditional on observing $Y(\mathcal{T}) = y(\mathcal{T})$, the predictive distribution of $X$ at $\mathcal{T}^{*} = (t^{*}_{1},\ldots,t^{*}_{m})$ is
\begin{equation}
  \label{eq:gp4}
  X(\mathcal{T}^{*})|\mathcal{T},Y(\mathcal{T}) = y(\mathcal{T}) \sim \mathcal{N}(\tilde{\mu}(\mathcal{T}^{*}),\tilde{\mbox{cov}}(\mathcal{T}^{*})), 
\end{equation}
where
\begin{equation}
  \label{eq:gp5}
  \begin{aligned}
    \tilde{\mu}(\mathcal{T}^{*}) &= \mu(\mathcal{T}^{*}) + \mathcal{K}_{\phi}(\mathcal{T}^{*},\mathcal{T})(\mathcal{K}_{\phi}(\mathcal{T},\mathcal{T}) + \sigma^{2} I_n)^{-1}(y(\mathcal{T}) - \mu(\mathcal{T})), \\
    \tilde{\mbox{cov}}(\mathcal{T}^{*}) &= \mathcal{K}_{\phi}(\mathcal{T}^{*},\mathcal{T}^{*}) - \mathcal{K}_{\phi}(\mathcal{T}^{*},\mathcal{T})(\mathcal{K}_{\phi}(\mathcal{T},\mathcal{T}) + \sigma^{2} I_n)^{-1}\mathcal{K}_{\phi}(\mathcal{T}, \mathcal{T}^{*}), \\
  \end{aligned}
\end{equation}
can be derived using the property of conditional multivariate Gaussian distribution. 

\paragraph{Derivative of Gaussian process} 
Let us now derive the first order derivative $\dot{X}(t)$ with respect to the input $t$. Since differentiation is a linear operator, the derivative of a Gaussian process is again a Gaussian process with some mean $\dot{\mu}$ (see Appendix of \citet{wenk2019gpode}), and the joint distribution of $X(t)$ and $\dot{X}(t)$ is 
\begin{equation}
  \label{eq:gp6}
  \left[\begin{array}{c} X(t) \\ \dot{X}(t)\end{array}\right] \sim \mathcal{GP}\left(\left[\begin{array}{c} \mu \\ \dot{\mu} \end{array}\right], \left[\begin{array}{cc} \mathcal{K}_{\phi} & \mathcal{K}_{\phi}^{\prime} \\ {}^{\prime}\mathcal{K}_{\phi} & \mathcal{K}_{\phi}^{\prime\prime} \end{array}\right] \right),
\end{equation}
where ${}^{\prime}\mathcal{K}_{\phi} = \frac{\partial}{\partial s}\mathcal{K}_{\phi}(s,t)$, $\mathcal{K}_{\phi}^{\prime} = \frac{\partial}{\partial t}\mathcal{K}_{\phi}(s,t)$, and $\mathcal{K}_{\phi}^{\prime\prime} = \frac{\partial}{\partial s \partial t}\mathcal{K}_{\phi}(s,t)$. Conditional on observing $X(\mathcal{T}) = x(\mathcal{T})$, the distribution of $\dot{X}(\mathcal{T})$ is 
\begin{equation}
  \label{eq:gp7}
  \dot{X}(\mathcal{T})|X(\mathcal{T}) = x(\mathcal{T}) \sim \mathcal{N}(\dot{\tilde{\mu}}(\mathcal{T}),\dot{\tilde{\mbox{cov}}}(\mathcal{T})),
\end{equation}
where
\begin{equation}
  \label{eq:gp8}
  \begin{aligned}
    \dot{\tilde{\mu}}(\mathcal{T}) &= \dot{\mu}(\mathcal{T}) + {}^{\prime}\mathcal{K}_{\phi}(\mathcal{T},\mathcal{T})\mathcal{K}_{\phi}(\mathcal{T},\mathcal{T})^{-1}(x(\mathcal{T}) - \mu(\mathcal{T})), \\
    \dot{\tilde{\mbox{cov}}}(\mathcal{T}) &= \mathcal{K}_{\phi}^{\prime\prime}(\mathcal{T},\mathcal{T}) - {}^{\prime}\mathcal{K}_{\phi}(\mathcal{T},\mathcal{T})\mathcal{K}_{\phi}(\mathcal{T},\mathcal{T})^{-1}\mathcal{K}_{\phi}^{\prime}(\mathcal{T}, \mathcal{T}), \\
  \end{aligned}
\end{equation}
are derived again by the property of conditional multivariate Gaussian distribution.

\paragraph{Hyperparameter Tuning}
We now discuss the tuning of the hyperparameters $\phi$. Noted from previous studies, the performance of the Gaussian process based gradient constrained methods rely heavily on the quality of the hyperparameters $\phi$. Following \citet{wenk2019gpode,wenk2020gpode,yang2020magi}, we employ the empirical Bayes method to choose the hyperparameters $\phi$ and the noise variance $\sigma^2$ that maximizes the marginal likelihood of the observations $y(\tau)$ defined in \eqref{eq:gp3}, that is to solve
\begin{equation}
  \label{eq:gp9}
  \arg\max_{\phi,\sigma^2}p_{\mathcal{N}}\bigg(y(\tau);\mu(\mathcal{T}),\mathcal{K}_{\phi}(\mathcal{T},\mathcal{T})+\sigma^{2} I_n\bigg)
\end{equation}
where $p_{\mathcal{N}}(\cdot;\mu,\Sigma)$ is the p.d.f. of multivariate Gaussian distribution with mean $\mu$ and variance $\Sigma$. If $\mu(\mathcal{T})$ is also not known apriori, we usually consider the prior mean function to be a constant function, i.e. $\mu(\cdot) = c$ for some unknown constant $c\in\mathbb{R}$. This is employed in our implementation, and we find $c$ by optimizing over $c$ along with $\phi$ and $\sigma^2$ in \eqref{eq:gp9}. In the case where $\mu(\cdot) = c$ is a constant function, then $\dot{\mu}(\cdot) = 0$ in \eqref{eq:gp6}.

\paragraph{Mat\'{e}rn covariance function} 
In this paper, we use the Mat\'{e}rn kernel with degree of freedom 2.01 that is suggested in \citet{yang2020magi}. The Mat\'{e}rn covariance function between any two points $t_1$ and $t_2$ with euclidean distance $d = \lVert t_1 - t_2\rVert_{2}$ is 
\begin{equation}
  \label{eq:matern}
  \mathcal{K}_{\nu}(t_{1},t_{2}) = \mathcal{K}_{\nu}(d) = \omega^2\frac{2^{1-\nu}}{\Gamma(\nu)}\bigg(\sqrt{2\nu}\frac{d}{\rho}\bigg)^{\nu}B_{\nu}\bigg(\sqrt{2\nu}\frac{d}{\rho}\bigg),
\end{equation}
where $\Gamma$ is the gamma function, $B_{\nu}$ is the modified Bessel function of the second kind, $\nu$ is the associated degree of freedom, $\omega$ is the variance parameter, and $\rho$ is the lengthscale parameter. The hyperparameters are $\phi = \{\omega,\rho\}$. Note that the modified Bessel function of the second kind satisfies the following recurrence relations:
\begin{equation}
  \label{eq:bessel1}
  \begin{aligned}
    &\frac{-2\nu}{r}B_{\nu}(r) = B_{\nu-1}(r) - B_{\nu+1}(r), \\
    &B^{\prime}_{\nu}(r) = -\frac{B_{\nu-1}(r) + B_{\nu+1}(r)}{2}, \\
  \end{aligned}
\end{equation}
and the following limit conditions:
\begin{equation}
  \label{eq:bessel2}
  \begin{aligned}
    &\lim_{r\to 0}B_{\nu}(r) = \infty, \\
    &\lim_{r\to 0}r^{\nu}B_{\nu}(r) = \frac{\Gamma(\nu)}{2^{1-\nu}}. \\
  \end{aligned}
\end{equation}
Thus, it follows that 
\begin{equation}
  \label{eq:bessel3}
  \lim_{d\to 0}K_{\nu}(d) = \omega^{2}\frac{2^{1-\nu}}{\Gamma(\nu)} \times \frac{\Gamma(\nu)}{2^{1-\nu}} = \omega^2.
\end{equation}
Let us now compute the partial derivative of Mat\'{e}rn covariance function with respect to $r = \sqrt{2\nu}{d}/\rho$,
\begin{equation}
  \label{eq:matern_1d_1}
  \begin{aligned}
  \frac{\partial \mathcal{K}_{\nu}}{\partial r} &= \omega^2\frac{2^{1-\nu}}{\Gamma(\nu)}\bigg\{\nu r^{\nu-1}B_{\nu}(r) + r^{\nu}B^{\prime}_{\nu}(r) \bigg\} \\
  &= \bigg\{\omega^2\frac{2^{1-\nu}}{\Gamma(\nu)}r^{\nu}B_{\nu}(r)\bigg\}\bigg\{\frac{\nu}{r}+\frac{B^{\prime}_{\nu}(r)}{B_{\nu}(r)}\bigg\} \\
  &= \mathcal{K}_{\nu}(d)\bigg\{\frac{\nu}{r}+\frac{B^{\prime}_{\nu}(r)}{B_{\nu}(r)}\bigg\} ,
  \end{aligned}
\end{equation}
with $\lim_{r\to 0} \partial \mathcal{K}_{\nu} / \partial r = 0$ (see Appendix~\ref{appendix:matern_limit_proof}). Given that we only consider one-dimensional inputs, the euclidean distance function can be simplified to be $d = |t_{1} - t_{2}|$, and thus we have
\begin{equation}
  \label{eq:matern_1d_2}
  \begin{aligned}
    \frac{\partial d}{\partial t_{1}} &= \frac{t_1 - t_2}{|t_1 - t_2|} \mbox{ if } t_1\neq t_2 \mbox{ and } 0 \mbox{ if } t_1 = t_2, \\
    \frac{\partial d}{\partial t_{2}} &= -\frac{t_1 - t_2}{|t_1 - t_2|} \mbox{ if } t_1\neq t_2 \mbox{ and } 0 \mbox{ if } t_1 = t_2. \\
  \end{aligned}
\end{equation}
One can recognize that $\partial d/\partial t_2 = -\partial d/\partial t_1$. By applying the chain rule, we can compute the first order partial derivative with respect to $t_{1}$ by
\begin{equation}
  \label{eq:matern_1d_3}
  \frac{\partial \mathcal{K}_{\nu}}{\partial t_{1}} = \frac{\partial \mathcal{K}_{\nu}}{\partial r} \frac{\partial r}{\partial d}\frac{\partial d}{\partial t_{1}},
\end{equation}
where $\partial r/\partial d = \sqrt{2\nu}/\rho$, and we can then compute $\partial \mathcal{K}_{\nu}/\partial t_{2} = -\partial \mathcal{K}_{\nu}/\partial t_{1}$. Next, let us compute the second order partial derivative with respect to both $t_{1}$ and $t_{2}$, that is
\begin{equation}
  \label{eq:matern_2d_1}
  \frac{\partial^2 \mathcal{K}_{\nu}}{\partial t_1 \partial t_2} = \frac{\partial}{\partial t_1}\bigg(\frac{\partial \mathcal{K}_{\nu}}{\partial d}\frac{\partial d}{\partial t_2}\bigg) = \frac{\partial^2 \mathcal{K}_{\nu}}{\partial d^2}\frac{\partial d}{\partial t_{1}}\frac{\partial d}{\partial t_{2}} + \frac{\partial \mathcal{K}_{\nu}}{\partial d}\frac{\partial^2 d}{\partial t_{1}\partial t_{2}} = -\frac{\partial^2 \mathcal{K}_{\nu}}{\partial d^2},
\end{equation}
by the fact that $\frac{\partial d}{\partial t_{1}}\frac{\partial d}{\partial t_{2}} = -(\frac{\partial d}{\partial t_{1}})^2 = -1$ and $\frac{\partial^2 d}{\partial t_{1}\partial t_{2}} = 0$. This leave us to compute $\partial^2 \mathcal{K}_{\nu}/\partial d^2$. Similarly, by using chain rule, we have
\begin{equation}
  \label{eq:matern_2d_2}
  \frac{\partial^2 \mathcal{K}_{\nu}}{\partial d^2} = \frac{\partial}{\partial d}\bigg(\frac{\partial \mathcal{K}_{\nu}}{\partial r}\frac{\partial r}{\partial d}\bigg) = \frac{\partial^2 \mathcal{K}_{\nu}}{\partial r^2}\frac{\partial r}{\partial d}\frac{\partial r}{\partial d} + \frac{\partial \mathcal{K}_{\nu}}{\partial r}\frac{\partial^2 r}{\partial d^2} = \frac{\partial^2 \mathcal{K}_{\nu}}{\partial r^2}\bigg(\frac{\partial r}{\partial d}\bigg)^2,
\end{equation}
where $\partial^2 r/\partial d^2 = 0$ from $r = \sqrt{2\nu}{d}/\rho$ and 
\begin{equation}
  \label{eq:matern_2d_3}
  \begin{aligned}
    \frac{\partial^2 \mathcal{K}_{\nu}}{\partial r^2} &= \omega^2\frac{2^{1-\nu}}{\Gamma(\nu)}\bigg\{\nu(\nu-1)r^{\nu-2}B_{\nu}(r) + 2\nu r^{\nu-1}B^{\prime}_{\nu}(r) + r^{\nu}B^{\prime\prime}_{\nu}(r) \bigg\} \\
    &= \bigg\{\omega^2\frac{2^{1-\nu}}{\Gamma(\nu)}r^{\nu}B_{\nu}(r)\bigg\}\bigg\{\frac{\nu(\nu-1)}{r^2}+\frac{2\nu}{r}\frac{B^{\prime}_{\nu}(r)}{B_{\nu}(r)}+\frac{B^{\prime\prime}_{\nu}(r)}{B_{\nu}(r)}\bigg\} \\
  &= \mathcal{K}_{\nu}(d)\bigg\{\frac{\nu(\nu-1)}{r^2}+\frac{2\nu}{r}\frac{B^{\prime}_{\nu}(r)}{B_{\nu}(r)}+\frac{B^{\prime\prime}_{\nu}(r)}{B_{\nu}(r)}\bigg\} ,
  \end{aligned}
\end{equation}
with $\lim_{r\to 0} \partial^2 \mathcal{K}_{\nu}/\partial r^2 = -\omega^2\frac{1}{2(\nu-1)}$ (see Appendix~\ref{appendix:matern_limit_proof}).

\section{Log Posterior}
\label{appendix:log_posterior}

Following the results in Appendix~\ref{appendix:gaussian_process}, we have 
\begin{equation}
  \label{eq:log_posterior1}
  X_{d}(\mathcal{T}) \sim \mathcal{N}\bigg(\mu_{d}(\mathcal{T}),C_{d}\bigg),
\end{equation}
where $C_{d} = \mathcal{K}_{d}(\mathcal{T},\mathcal{T})$, and 
\begin{equation}
  \label{eq:log_posterior2}
  \dot{X}_{d}(\mathcal{T})|X_{d}(\mathcal{T}) = x_{d}(\mathcal{T}) \sim \mathcal{N}\bigg(\dot{\tilde{\mu}}_{d}(\mathcal{T}),K_{d}\bigg),
\end{equation}
where 
\begin{equation}
  \label{eq:log_posterior3}
  \begin{aligned}
    \dot{\tilde{\mu}}_{d}(\mathcal{T}) &= \dot{\mu}_{d}(\mathcal{T}) + {}^{\prime}\mathcal{K}_{d}(\mathcal{T},\mathcal{T})\mathcal{K}_{d}(\mathcal{T},\mathcal{T})^{-1}(x_{d}(\mathcal{T}) - \mu_{d}(\mathcal{T})), \\
    K_{d} &= \mathcal{K}_{d}^{\prime\prime}(\mathcal{T},\mathcal{T}) - {}^{\prime}\mathcal{K}_{d}(\mathcal{T},\mathcal{T})\mathcal{K}_{d}(\mathcal{T},\mathcal{T})^{-1}\mathcal{K}_{d}^{\prime}(\mathcal{T}, \mathcal{T}), \\
  \end{aligned}
\end{equation}
follows from \eqref{eq:gp8}. Thus,
\begin{equation}
  \label{eq:log_posterior4}
  \begin{aligned}
    & p(W_{\mathcal{T}} = 0|X(\mathcal{T}) = x(\mathcal{T}), \Theta = \theta) \\
    = &\prod_{d=1}^{D} p\bigg(\dot{X}_{d}(\mathcal{T}) = \{f(x(\mathcal{T}),\theta,t_{\mathcal{T}})\}_{d} |X(\mathcal{T}) = x(\mathcal{T}), \Theta = \theta\bigg) \\
    = &\prod_{d=1}^{D} p_{\mathcal{N}}\bigg(\{f(x(\mathcal{T}),\theta,t_{\mathcal{T}})\}_{d}; \dot{\tilde{\mu}}_{d}(\mathcal{T}),K_{d}\bigg), \\
  \end{aligned}
\end{equation}
where $p_{\mathcal{N}}(\cdot;\mu,\Sigma)$ is the p.d.f. of multivariate Gaussian distribution with mean $\mu$ and variance $\Sigma$. Thus, the log posterior distribution function is 
\begin{equation}
  \label{eq:log_posterior5}
  \begin{aligned}
  & \log p\{\Theta=\theta,X(\mathcal{T})=x(\mathcal{T})|W_{\mathcal{T}}=0,Y(\tau)=y(\tau)\} \\
  =\; & \text{Const.} + \log\pi_{\Theta}(\theta) - \frac{1}{2}\sum_{d=1}^{D}\bigg\{\log|C_{d}| + \lVert x_{d}(\mathcal{T}) - \mu_{d}(\mathcal{T})\rVert_{C_{d}^{-1}}^{2} + \\
  & N_{d}\log(\sigma_{d}^2) + \lVert y_{d}(\tau_{d}) - x_{d}(\tau_{d})\rVert_{\sigma_{d}^{-2}I_{N_{d}}}^{2} + \log|K_{d}| + \lVert \{f(x(\mathcal{T}),\theta,t_{\mathcal{T}})\}_{d} - \dot{\tilde{\mu}}_{d}(\mathcal{T}) \rVert_{K_{d}^{-1}} \bigg\}
  \end{aligned}
\end{equation}
where $\lVert v\rVert_{A}^{2} = v^{T}A v$ and $|A|$ is the determinant of $A$.

\section{Additional Simulation Results}
\label{appendix:additional_simulation_results}

\subsection{Benchmark Real World Systems}
\label{appendix:benchmark_real_world_systems}

\begin{figure}[t!]
  \centering
  \begin{subfigure}{0.32\textwidth}
    \centering
    \includegraphics[width=0.95\textwidth]{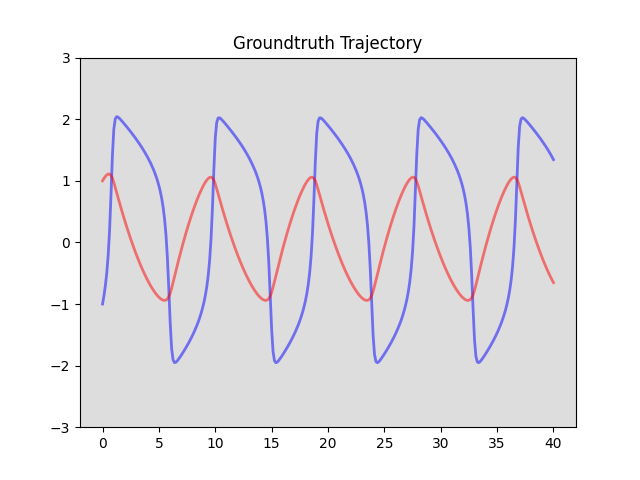}
    \caption{FN}
  \end{subfigure}
  \begin{subfigure}{0.32\textwidth}
    \centering
    \includegraphics[width=0.95\textwidth]{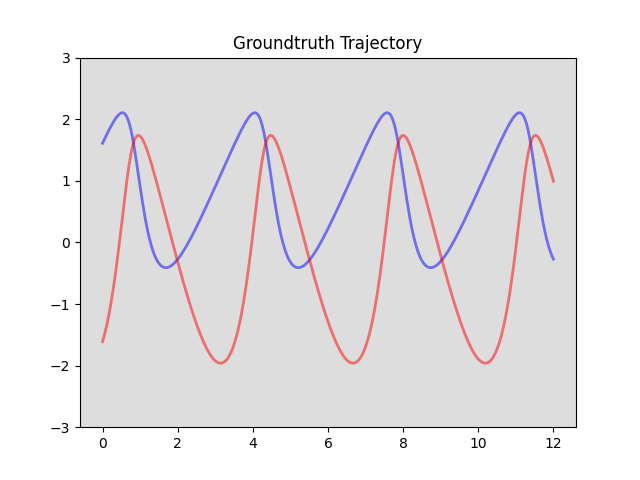}
    \caption{LV (log-transformed)}
  \end{subfigure}
  \begin{subfigure}{0.32\textwidth}
    \centering
    \includegraphics[width=0.95\textwidth]{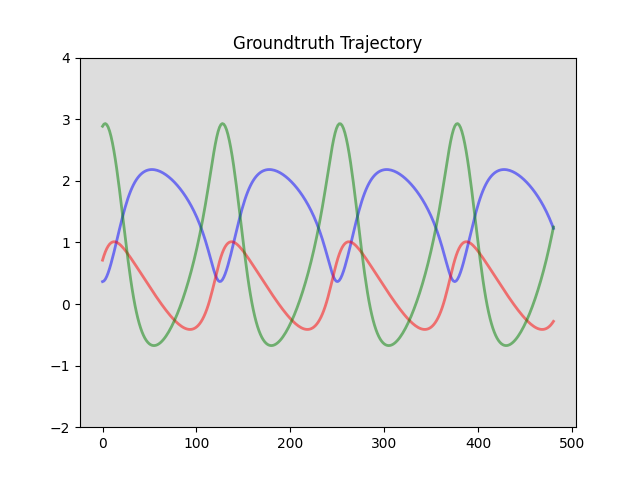}
    \caption{Hes1 (log-transformed)}
  \end{subfigure}
  \caption{Groundtruth trajectory obtained using numerical integration.}
  \label{fig:groundtruth}
\end{figure}

In this paper, we consider the following three benchmark real world systems:

\begin{itemize}
  \item The FitzHugh-Nagumo (FN) system was introduced by \citet{fitzhugh1961fn} and \citet{nagumo1962fn} for modeling the activation of an excitable system such as neuron. It is a two-component system determined by the following ODEs,
  \begin{equation}
    \label{eq:fn}
    \left\{\begin{array}{ll}
      \dot{x}_{1} = c(x_{1} - x_{1}^{3}/3 + x_{2}) \\ 
      \dot{x}_{2} = -(x_{1} - a + b x_{2}) / c \\
    \end{array}\right.
  \end{equation}
  where $a = 0.2$, $b = 0.2$, $c = 3$, and $x(0) = (-1,1)$. The groundtruth trajectory from $t=0$ to $t=40$ obtained via numerical integration is presented in the left panel of Figure~\ref{fig:groundtruth}.
  
  \item The Lotka-Volterra (LV) system was introduced by \citet{lotka1978lv} for modeling the dynamics of the predator-prey interaction. It is a two-component system determined by the following ODEs,
  \begin{equation}
    \label{eq:lv}
    \left\{\begin{array}{ll}
      \dot{x}_{1} = a x_{1} - b x_{1}x_{2} \\ 
      \dot{x}_{2} = c x_{1}x_{2} - d x_{2} \\
    \end{array}\right.
  \end{equation}
  where $a = 1.5$, $b = 1$, $c = 1$, $d = 3$, and $x(0) = (5, 0.2)$. Given that both $x_{1}$ and $x_{2}$ are always strictly positive, we consider the log-transformation of the system in this paper. The groundtruth trajectory from $t=0$ to $t=12$ obtained via numerical integration is presented in the middle panel of Figure~\ref{fig:groundtruth}.
  
  \item The Hes1 system was introduced by \citet{hirata2002hes1} for modeling the oscillation dynamic of Hes1 protein level ($x_{1}$) and Hes1 mRNA level ($x_{2}$) under the influence of a Hes1-interacting factor ($x_{3}$). It is a three-component system determined by the following ODEs,  
  \begin{equation}
    \label{eq:hes1}
    \left\{\begin{array}{ll}
      \dot{x}_{1} = -a x_{1}x_{3} + b x_{2} - c x_{1} \\ 
      \dot{x}_{2} = -d x_{2} + e / (1+x_{1}^2) \\ 
      \dot{x}_{3} = -a x_{1}x_{3} + f/(1+x_{1}^2) - g x_{3}\\ 
    \end{array}\right.
  \end{equation}
  where $a = 0.022$, $b = 0.3$, $c = 0.031$, $d = 0.028$, $e = 0.5$, $f = 20$, $g = 0.3$, and $x(0) = (1.438575, 2.037488, 17.90385)$. Similarly, given that $x_{1}$, $x_{2}$, and $x_{3}$ are always strictly positive, we consider the log-transformation of the system in this paper. The groundtruth trajectory from $t=0$ to $t=480$ obtained via numerical integration is presented in the right panel of Figure~\ref{fig:groundtruth}.
\end{itemize}

\begin{table}
  \caption{Means and standard deviations of trajectory RMSEs over 100 runs with full observations.}
  \label{tab:trajectory_rmse_magi-x}
  \centering
  \resizebox{\columnwidth}{!}{%
  \begin{tabular}{lllccccccc}
    \toprule
    & & & \multicolumn{2}{c}{FN} & \multicolumn{2}{c}{LV} & \multicolumn{3}{c}{Hes1} \\
    \cmidrule(lr){4-5}\cmidrule(lr){6-7}\cmidrule(lr){8-10}
    Phase & Model & Trajectory & $x_1$ & $x_2$ & $x_1$ & $x_2$ & $x_1$ & $x_2$ & $x_3$ \\
    \midrule
    \multirow{4}{5em}{Fitting} & \multirow{2}{5em}{MAGI-X\\(512)} & Reconstructed & $0.21 \pm 0.02$ & $0.05 \pm 0.01$ & $0.06 \pm 0.01$ & $0.07 \pm 0.03$ & $0.12 \pm 0.05$ & $0.06 \pm 0.02$ & $0.13 \pm 0.03$ \\
    & & Inferred & $0.21 \pm 0.02$ & $0.05 \pm 0.01$ & $0.05 \pm 0.01$ & $0.06 \pm 0.01$ & $0.11 \pm 0.05$ & $0.06 \pm 0.01$ & $0.12 \pm 0.04$ \\
    \cmidrule{2-10}
    & \multirow{2}{5em}{MAGI-X\\(32,32,32)} & Reconstructed & $0.38 \pm 0.10$ & $0.06 \pm 0.02$ & $0.05 \pm 0.02$ & $0.07 \pm 0.03$ & $0.08 \pm 0.03$ & $0.06 \pm 0.02$ & $0.13 \pm 0.05$ \\
    & & Inferred & $0.37 \pm 0.10$ & $0.05 \pm 0.01$ & $0.05 \pm 0.01$ & $0.05 \pm 0.01$ & $0.08 \pm 0.03$ & $0.06 \pm 0.02$ & $0.12 \pm 0.05$  \\
    
    \midrule
    
    \multirow{4}{5em}{Forecasting} & \multirow{2}{5em}{MAGI-X\\(512)} & Reconstructed & $0.21 \pm 0.03$ & $0.06 \pm 0.02$ & $0.12 \pm 0.10$ & $0.18 \pm 0.24$ & $0.25 \pm 0.14$ & $0.13 \pm 0.07$ & $0.30 \pm 0.17$  \\
    & & Inferred & $0.21 \pm 0.03$ & $0.06 \pm 0.02$ & $0.09 \pm 0.04$ & $0.12 \pm 0.06$ & $0.24 \pm 0.14$ & $0.13 \pm 0.06$ & $0.30 \pm 0.16$ \\
    \cmidrule{2-10}
    & \multirow{2}{5em}{MAGI-X\\(32,32,32)} & Reconstructed & $0.47 \pm 0.24$ & $0.14 \pm 0.13$ & $0.14 \pm 0.15$ & $0.21 \pm 0.29$ & $0.21 \pm 0.15$ & $0.13 \pm 0.09$ & $0.35 \pm 0.27$  \\
    & & Inferred & $0.42 \pm 0.13$ & $0.11 \pm 0.06$ & $0.10 \pm 0.06$ & $0.13 \pm 0.09$ & $0.21 \pm 0.16$ & $0.13 \pm 0.10$ & $0.36 \pm 0.29$ \\
    \bottomrule
  \end{tabular}
  }
\end{table}

MAGI-X utilizes the neural networks for modeling the unknown derivative function $f$. It would be interesting to investigate how different network structures might affect the performance of the modeling accuracy. The \textit{expressive power} of the neural networks has been studied theoretically since the late 80's, starting with the pioneering work of \citet{cybenko1989nn,hornik1989nn,barron1994nn} that state the universal approximation theorem: the neural networks can approximate arbitrary continuous function on a compact domain to any desired accuracy. The universal approximation theorem holds even for the shallow neural networks with single hidden layer, but this comes in the expense of requiring a very wide hidden layer with size being exponential of the input dimension even for approximating the simple multiplication \citep{lin2017nn}. Such inefficiency can be addressed by the use of deeper networks in some of the recent theoretical investigations: deep networks requires exponentially smaller number of neurons to achieve same approximation accuracy by the shallow networks \citep{eldan2016nn,telgarsky2016nn,rolnick2018nn}. For the deep networks to be universal approximator, they have to satisfy certain lower bound for the network width \citep{lu2017nn,park2021nn}. 

Inspired by the aforementioned theoretical works, we consider both wide shallow network (one hidden layers with 512 nodes) and deeper network (three hidden layers each with 32 nodes) for the modeling of $f$. Table~\ref{tab:trajectory_rmse_magi-x} shows the means and standard deviations of the trajectory RMSEs over 100 simulated runs with full observations under the same setting discussed in the paper. We can see that the results are comparable but the wide shallow network (one hidden layers with 512 nodes) performs better on the two-component systems (FN and LV). Some numerical tricks such as batch normalization might be needed for yielding better and more robust performance from the deeper networks. This is left for future investigation. 

\begin{figure}[t!]
  \centering
  \begin{subfigure}{0.32\textwidth}
    \centering
    \includegraphics[width=0.95\textwidth]{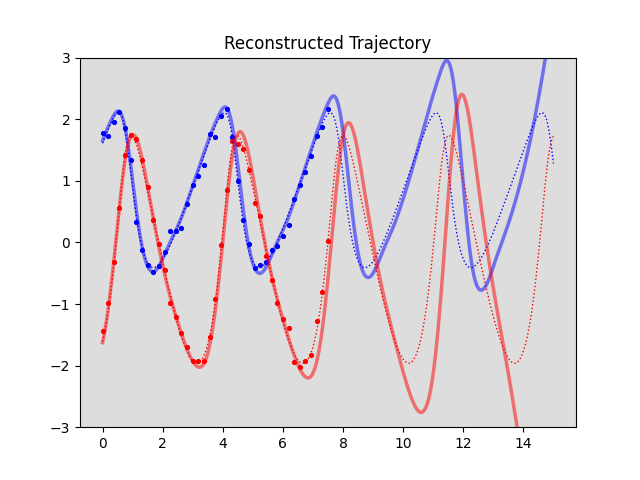}
    \caption{Seed 1}
  \end{subfigure}
  \begin{subfigure}{0.32\textwidth}
    \centering
    \includegraphics[width=0.95\textwidth]{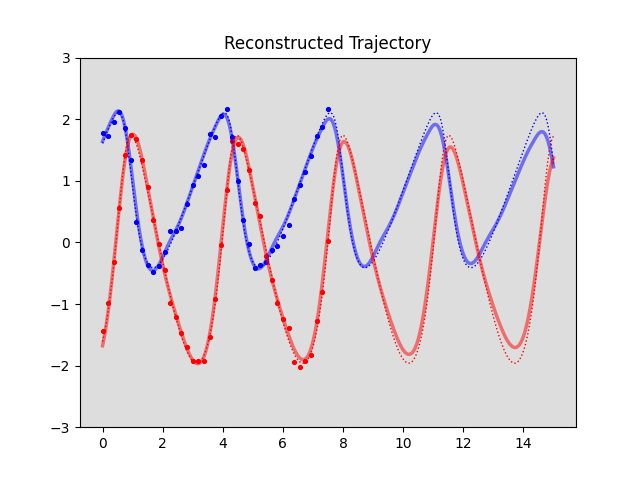}
    \caption{Seed 2}
  \end{subfigure}
  \caption{Comparison of the reconstructed trajectory (solid line) to the groundtruth trajectory (dotted lines) after applying MAGI-X on the same set of 41-point noisy data (circles) from the LV system but with two different initialized random seed for the optimization procedure.}
  \label{fig:lv_discussion}
\end{figure}

However, we do observe some large standard deviations in the trajectory RMSEs from Table~\ref{tab:trajectory_rmse_magi-x}, e.g., the reconstructed trajectory RMSE of $x_{2}$ for the LV system with one hidden layer of 512 nodes neural network. This large standard deviation is due to one divergence case where the small errors accumulate in the trajectory propagation via numerical integration in the forecasting phase, and eventually push the trajectory to unseen domain where we do not have data for the neural network to learn the correct behavior (left panel of Figure~\ref{fig:lv_discussion}). However, using a different initialized random seed for the optimization solves the problem (right panel of Figure~\ref{fig:lv_discussion}). Thus, in real world usage, we can actually restart MAGI-X with multiple random seeds and select the best one to avoid the possible divergence, and this could still result in faster runtime than NPODE and Neural ODE given the computational saving achieved by MAGI-X.

Nevertheless, improvements to the optimization procedure are still encouraged. One promising solution is to provide a more variety of trajectories $x(\mathcal{T})$ but are still within some credible interval of the true trajectory for training $\theta$, so the model has data for learning about auto-correction once the trajectory slightly deviates. Thus, one future direction is to replace SGD for the update of $x(\mathcal{T})$ by the stochastic gradient Langevin dynamics \citep{welling2011sgld} or stochastic gradient Markov Chain Monte Carlo \citep{ma2015sgmcmc} that are both stochastic gradient based samplers with carefully injected noise such that (i) ``nosier" samples are available for the training of $\theta$, and (ii) the samples can represent the target posterior, so we can do uncertainty quantification for the inferred trajectory $x(\mathcal{T})$, where the current MAGI-X does not support.

\subsection{Synthetic High Dimensional Systems}
\label{subsec:synthetic_high_dimensional_systems}

To study the performance of MAGI-X in the large system setting, we consider the following ODEs derived from the Hamiltonian system $H(q,p) = \frac{1}{2}p^{T}p + q^{T}q$,
\begin{equation}
  \label{eq:large_system}
  \left\{\begin{array}{l} \dot{p} = -\frac{\partial H}{\partial q} = -2q \\ \dot{q} = \frac{\partial H}{\partial p} = p \\ \end{array}\right.
\end{equation}
where $q$ and $p$ are both $n$-dimensional vector, and thus the system has total of $D=2n$ components. The initial value of the system is simulated randomly from the standard Gaussian distribution. We consider the ODEs system with $n=5,10,20$, which corresponds to the number of components $D=10,20,40$ for demonstration. Similar to the data generation procedure in the three real world systems, we again inject $0.1^2$ variance i.i.d. Gaussian random noise to the groundtruth trajectory to obtain the 41-point noisy observations for each component. MAGI-X is then ran for 5{,}000 iterations with the default learning rate.

\begin{table}[t!]
  \caption{Means and standard deviations of trajetory RMSEs over 100 runs on the synthetic large system example after excluding the divergence cases.}
  \label{tab:large_system}
  \centering
  \begin{tabular}{ccccccc}
    \toprule
    & & \multicolumn{2}{c}{Reconstructed} & \multicolumn{2}{c}{Inferred} \\
    \cmidrule(lr){3-4}\cmidrule(lr){5-6}
    Dimension & No. Divergence & Imputation & Forecast & Imputation & Forecast \\
    \midrule
    10 & 4 & $0.08 \pm 0.05$ & $0.38 \pm 0.40$ & $0.06 \pm 0.01$ & $0.30 \pm 0.25$ \\
    20 & 10 & $0.07 \pm 0.02$ & $0.41 \pm 0.35$ & $0.06 \pm 0.01$ & $0.36 \pm 0.28$ \\
    40 & 11 & $0.07 \pm 0.01$ & $0.46 \pm 0.29$ & $0.07 \pm 0.01$ & $0.46 \pm 0.30$ \\
    \bottomrule
  \end{tabular}
\end{table}

We have presented the computational time in the main paper: runtime scales linearly in the component dimensions. In the appendix we focus on the accuracy performance of MAGI-X. Given that our optimization could still once in a while stuck at local optimum and yield a divergent forecasting trend as in the LV example (Figure~\ref{fig:lv_discussion}), we would expect the same limitation in the larger system example, and thus we include the number of divergence case as another measure for evaluating the performance. We define the divergence to be the case if any component's forecasting trajectory RMSE is greater than 5, which indicates that the forecasting trend is far away from the observation domain of \eqref{eq:large_system}. Table~\ref{tab:large_system} shows the trajectory RMSEs excluding the divergence cases. We can see that the recovery of the true dynamic is accurate for the imputation phase, but the performance on the forecasting is not as good. Moreover, we do observe more divergence case as the number of components increase, which is expected as the higher dimensional problem is harder. However, even on the 40-component system, we only observe about 10\% divergence case, showing that a few restarts would be sufficient to avoid the divergence problem, which is still a significant computational saving from the existing methods (e.g. runtime of MAGI-X on the 40-component system is 270 seconds, while runtime of Neural ODE on the 3-component system is more than 500 seconds).

\section{Implementation Details}
\label{appendix:implementation_details}

We finally discuss some implementation details that we employ in implementation to avoid numerical instability. 

\subsection{Time standardization} 
Given that the FN, LV, and Hes1 examples share similar magnitude in their component values but have very different time ranges (Figure~\ref{fig:groundtruth}), some time standardization might help improve the robustness of the algorithm. Though different time unit would give theoretically equivalent system, we do observe that NPODE of \citet{heinonen2018odegp} yields very bad performance on the Hes1 example, but such problem does not exist for the FN and LV systems (time range in the scale of 10s). Same phenomenon is also observed in MAGI-X \textit{without} the time standardization. Given that both NPODE and MAGI-X rely on gradient update for parameters optimization, the performance is sensitive to the learning rate. Thus, the poor performance on the Hes1 example suggests that the learning rate chosen in NPODE and MAGI-X is robust for system with time scale of 10s, but it could be too large for system with time scales of 100s if the component values have similar magnitude. This leads to the need of time standardization. From simulation studies on the three benchmark problems with the specified learning rate, we suggest the following standardization scheme: standardize the data such that the distance between any two nearby time points is 0.05. Given that $\mathcal{T}$ is the 161 equal spaced-out time points from fitting phase, we are essentially standardizing the fitting phase time range to 8 for all systems. What we propose now is an engineering solution, and a better standardization procedure will be investigated in future works.

\subsection{Cholesky decomposition}
Optimizing over $x(\mathcal{T})$ directly in the objective function of MAGI-X is usually not numerically preferred as the entries of $x_{d}(\mathcal{T})$ are supposed to be correlated by the GP prior. Now consider $U(\mathcal{T}) = (U_{1}(\mathcal{T}),\ldots,U_{D}(\mathcal{T}))$ where 

\begin{equation}
  \label{eq:cholesky}
  U_{d}(\mathcal{T}) = L_{C_{d}}^{-1}(X_{d}(\mathcal{T}) - \mu_{d}(\mathcal{T})) \quad \Leftrightarrow \quad X_{d}(\mathcal{T}) = \mu_{d}(\mathcal{T}) + L_{C_{d}}U_{d}(\mathcal{T}).
\end{equation}

$L_{C_{d}}$ is the Cholesky decomposition of $C_{d} = \mathcal{K}_{d}(\mathcal{T},\mathcal{T})$, i.e., $C_{d} = L_{C_{d}} L_{C_{d}}^{T}$. We can see that $U_{d}\sim\mathcal{N}(0,I_{|\mathcal{T}|})$, where all the entries are independent. We make the change of variable from $X(\mathcal{T})$ to $U(\mathcal{T})$ in the objective function, and optimize over $u(\mathcal{T}) = (u_{1}(\mathcal{T}),\ldots,u_{D}(\mathcal{T})$ where $u_{d}(\mathcal{T}) = L_{C_{d}}^{-1}(x_{d}(\mathcal{T}) - \mu_{d}(\mathcal{T}))$ in the actual implementation. This leads to more robust performance when gradient update is employed.

\section{Proof for the limit of Mat\'{e}rn covariance function's derivative at 0}
\label{appendix:matern_limit_proof}

\textbf{Claim 1:} $\lim_{r\to 0}\frac{\partial \mathcal{K}_{\nu}}{\partial r} = 0$.

\begin{proof}

Follow from \eqref{eq:matern_1d_1} and apply the recurrence relations \eqref{eq:bessel1} and the limit conditions \eqref{eq:bessel2} of the modified Bessel function of the second kinds, we have

\begin{equation}
  \label{eq:matern_limit_1d_1}
  \begin{footnotesize}
    \begin{aligned}
      &\lim_{r\to 0}\frac{\partial \mathcal{K}_{\nu}}{\partial r} \\
      =& \omega^2 \frac{2^{1-\nu}}{\Gamma(\nu)} \lim_{r\to 0} \bigg\{\frac{\nu}{r}r^{\nu}B_{\nu}(r) + r^{\nu}B^{\prime}_{\nu}(r) \bigg\} \\
      =& \omega^2 \frac{2^{1-\nu}}{\Gamma(\nu)} \lim_{r\to 0} \bigg\{\frac{\nu}{r}r^{\nu}B_{\nu}(r) + r^{\nu}\bigg(-\frac{B_{\nu-1}(r) + B_{\nu+1}(r)}{2}\bigg) \bigg\} \\
      =& \omega^2 \frac{2^{1-\nu}}{\Gamma(\nu)} \lim_{r\to 0}\bigg\{\frac{\nu r^{\nu}B_{\nu}(r) - \frac{1}{2}r^{\nu+1}B_{\nu+1}(r)}{r} - \frac{1}{2}r^{\nu}B_{\nu-1}(r)\bigg\} \\
      =& \omega^2 \frac{2^{1-\nu}}{\Gamma(\nu)} \lim_{r\to 0}\frac{\nu r^{\nu}B_{\nu}(r) - \frac{1}{2}r^{\nu+1}B_{\nu+1}(r)}{r} \\
      =& \omega^2 \frac{2^{1-\nu}}{\Gamma(\nu)} \lim_{r\to 0}\frac{\nu r^{\nu}\{-\frac{r}{2\nu}(B_{\nu-1}(r)-B_{\nu+1}(r))\} - \frac{1}{2}r^{\nu+1}B_{\nu+1}(r)}{r} \\
      =& \omega^2 \frac{2^{1-\nu}}{\Gamma(\nu)} \lim_{r\to 0}\frac{-\frac{1}{2}r^{\nu+1}B_{\nu-1}(r)}{r} \\
      =& \omega^2 \frac{2^{1-\nu}}{\Gamma(\nu)} \lim_{r\to 0}-\frac{1}{2}r^{\nu}B_{\nu-1}(r) \\
      =& \omega^2 \frac{2^{1-\nu}}{\Gamma(\nu)} \lim_{r\to 0}r\bigg(-\frac{1}{2}r^{\nu-1}B_{\nu-1}(r)\bigg) \\
      =& 0 \\
    \end{aligned}
  \end{footnotesize}
\end{equation}

\end{proof}

\textbf{Claim 2:} $\lim_{r\to 0}\frac{\partial^2 \mathcal{K}_{\nu}}{\partial r^2} = -\omega^2\frac{1}{2(\nu-1)}$.

\begin{proof}

Recursively applying the recurrence relations \eqref{eq:bessel1}, we can compute that

\begin{equation}
  \label{eq:matern_limit_2d_1}
  \begin{footnotesize}
    \begin{aligned}
    B^{\prime\prime}(r) = \frac{1}{4}\bigg(B_{\nu-2}(r) + 2B_{\nu}(r) + B_{\nu+2}(r)\bigg) .
    \end{aligned}
  \end{footnotesize}
\end{equation}

Thus, applying the recurrence relations \eqref{eq:bessel1} and the limit conditions \eqref{eq:bessel2} to \eqref{eq:matern_2d_3} as $r\to 0$, we have 

\begin{equation}
  \label{eq:matern_limit_2d_2}
  \begin{footnotesize}
  \begin{aligned}
    &\lim_{r\to 0}\frac{\partial^2\mathcal{K}_{\nu}}{\partial r^2} \\
    =& \omega^2 \frac{2^{1-\nu}}{\Gamma(\nu)} \lim_{r\to 0} \bigg\{\frac{\nu(\nu-1)}{r^2}r^{\nu}B_{\nu}(r) + \frac{2\nu}{r}r^{\nu}B^{\prime}(\nu) + r^{\nu}B^{\prime\prime}_{\nu}(r) \bigg\} \\
    =&\omega^2 \frac{2^{1-\nu}}{\Gamma(\nu)} \lim_{r\to 0}\bigg\{\frac{\nu(\nu-1)}{r^2}r^{\nu}B_{\nu}(r) + \frac{2\nu}{r}r^{\nu}\bigg(-\frac{B_{\nu-1}(r) + B_{\nu+1}(r)}{2}\bigg) + \\
    &r^{\nu}\frac{1}{4}\bigg(B_{\nu-2}(r) + 2B_{\nu}(r) + B_{\nu+2}(r)\bigg) \bigg\} \\
    =&\omega^2 \frac{2^{1-\nu}}{\Gamma(\nu)} \lim_{r\to 0}\bigg\{\frac{\nu(\nu-1)}{r^2}r^{\nu}B_{\nu}(r) - \nu r^{\nu-1}B_{\nu-1}(r) - \frac{\nu}{r^2}r^{\nu+1}B_{\nu+1}(r) + \\
    &\frac{r^2}{4}r^{\nu-2}B_{\nu-2}(r) + \frac{1}{2}r^{\nu}B_{\nu}(r) + \frac{1}{r^2}\frac{1}{4}r^{\nu+2}B_{\nu+2}(r) \bigg\} \\
    =&\omega^2 \frac{2^{1-\nu}}{\Gamma(\nu)} \lim_{r\to 0}\bigg\{-r^{\nu-1}B_{\nu-1}(r) + \bigg(-(\nu-1)r^{\nu-1}B_{\nu-1}(r) + \frac{1}{2}r^{\nu}B_{\nu}(r)\bigg) + \\
    &\frac{\nu(\nu-1)r^{\nu}B_{\nu}(r) - \nu r^{\nu+1}B_{\nu+1}(r) + \frac{1}{4}r^{\nu+2}B_{\nu+2}(r)}{r^2} \bigg\} \\
    =& \omega^2 \frac{2^{1-\nu}}{\Gamma(\nu)}\lim_{r\to 0}\bigg\{-r^{\nu-1}B_{\nu-1}(r)\bigg\} \\
    =& \omega^2 \frac{2^{1-\nu}}{\Gamma(\nu)}\bigg\{-\frac{\Gamma(\nu-1)}{2^{1-(\nu-1)}}\bigg\} \\
    =& -\omega^2 \frac{1}{2(\nu-1)}
  \end{aligned}
  \end{footnotesize}
\end{equation}

where 

\begin{equation}
  \label{eq:matern_limit_2d_3}
  \begin{footnotesize}
  \begin{aligned}
    \lim_{r\to 0} \bigg(-(\nu-1)r^{\nu-1}B_{\nu-1}(r) + \frac{1}{2}r^{\nu}B_{\nu}(r)\bigg) = -(\nu-1)\frac{\Gamma(\nu-1)}{2^{1-(\nu-1)}} + \frac{1}{2}\cdot\frac{\Gamma(\nu)}{2^{1-\nu}} = 0
  \end{aligned}
  \end{footnotesize}
\end{equation}

and 

\begin{equation}
  \label{eq:matern_limit_2d_4}
  \begin{footnotesize}
  \begin{aligned}
    & \lim_{r\to 0} \frac{\nu(\nu-1)r^{\nu}B_{\nu}(r) - \nu r^{\nu+1}B_{\nu+1}(r) + \frac{1}{4}r^{\nu+2}B_{\nu+2}(r)}{r^2} \\
    =& \lim_{r\to 0} \frac{\nu(\nu-1)r^{\nu}\{-\frac{r}{2\nu}(B_{\nu-1}(r)-B_{\nu+1}(r))\} - \nu r^{\nu+1}B_{\nu+1}(r) + \frac{1}{4}r^{\nu+2}B_{\nu+2}(r)}{r^2} \\
    =& \lim_{r\to 0} \frac{-\frac{\nu-1}{2}r^{\nu+1}B_{\nu-1}(r)+\frac{\nu-1}{2}r^{\nu+1}B_{\nu+1}(r) - \nu r^{\nu+1}B_{\nu+1}(r) + \frac{1}{4}r^{\nu+2}B_{\nu+2}(r)}{r^2} \\
    =& \lim_{r\to 0} \frac{-\frac{\nu-1}{2}r^{\nu+1}\{-\frac{r}{2(\nu-1)}(B_{\nu-2}(r)-B_{\nu}(r))\}-\frac{\nu+1}{2}r^{\nu+1}B_{\nu+1}(r) + \frac{1}{4}r^{\nu+2}B_{\nu+2}(r)}{r^2} \\
    =& \lim_{r\to 0} \frac{-\frac{1}{4}r^{\nu+2}B_{\nu}(r) - \frac{\nu+1}{2}r^{\nu+1}B_{\nu+1}(r) + \frac{1}{4}r^{\nu+2}B_{\nu+2}(r)}{r^2} \\
    =& \lim_{r\to 0} \frac{-\frac{\nu+1}{2}r^{\nu+1}B_{\nu+1}(r) - \frac{1}{4}r^{\nu+2}(B_{\nu}(r) - B_{\nu+2}(r))}{r^2} \\
    =& \lim_{r\to 0} \frac{-\frac{\nu+1}{2}r^{\nu+1}B_{\nu+1}(r) - \frac{1}{4}r^{\nu+2}\{-\frac{2(\nu+1)}{r}B_{\nu+1}(r)\}}{r^2} \\
    =& \lim_{r\to 0} \frac{-\frac{\nu+1}{2}r^{\nu+1}B_{\nu+1}(r) + \frac{\nu+1}{2}r^{\nu+1}B_{\nu+1}(r)}{r^2} \\
    =& \lim_{r\to 0} \frac{0}{r^2} \\
    =& 0
  \end{aligned}
  \end{footnotesize}
\end{equation}

and the last equality follows from the L'H{\^o}spital's Rule.

\end{proof}

\end{document}